\pdfoutput=1
\documentclass[11pt]{article}

\usepackage{style/acl}

\usepackage{times}
\usepackage{latexsym}
\usepackage{enumitem}
\setlist{nosep}
\usepackage{tabularx}
\usepackage{microtype}
\usepackage{graphicx}
\usepackage{booktabs} % for professional tables
\usepackage{amsmath}
\usepackage{amsfonts}
\usepackage{subcaption}
\usepackage[all]{nowidow}
\definecolor{grayish}{rgb}{0.95, 0.95, 0.96}
\definecolor{corn}{RGB}{233, 210, 135}

\newcommand{\cicero}{Cicero}

\newcommand{\amr}[1]{\texttt{#1}}
\usepackage[T1]{fontenc}
\usepackage[utf8]{inputenc}

\usepackage{microtype}

\usepackage{xcolor}
\usepackage{hyperref}
\usepackage{inconsolata}
\newcommand{\email}[1]{\textcolor{blue}{\small\texttt{\href{mailto:#1}{#1}}}}

\title{More Victories, Less Cooperation: Assessing \cicero{}'s Diplomacy Play}

\author{Wichayaporn Wongkamjan$^{1}$ \hspace{0.5cm} Feng Gu$^{1}$ \hspace{0.5cm} Yanze Wang$^{4}$ \hspace{0.5cm} Ulf Hermjakob$^{4}$ \\
{\bf Jonathan May}$^{4}$ \hspace{0.5cm} {\bf Brandon M. Stewart}$^{2}$ \hspace{0.5cm} {\bf Jonathan K. Kummerfeld}$^{3}$ \\
{\bf Denis Peskoff}$^{2}$ \hspace{0.5cm} {\bf Jordan Lee Boyd-Graber$^{1}$}\\
$^{1}$University of Maryland \hspace{0.5cm} $^{2}$Princeton University \hspace{0.5cm} $^{3}$University of Sydney \\ $^{4}$Information Sciences Institute, University of Southern California \\
\email{\{wwongkam,fgu1\}@umd.edu} \hspace{0.2cm} \hspace{0.2cm} 
\email{\{yanzewan,ulf,jonmay\}@isi.edu} \\
\email{bms4@princeton.edu} \hspace{0.2cm}
\email{jonathan.kummerfeld@sydney.edu.au} \hspace{0.2cm} \\
\email{dp2896@princeton.edu} \hspace{0.2cm}
\email{jbg@umiacs.umd.edu} \hspace{0.2cm}
}

\newif\ifcomment\commentfalse
\usepackage[a-1b]{pdfx}

\usepackage{framed}
\usepackage{mdwlist}
\usepackage{siunitx}
\usepackage{latexsym}
\usepackage{colortbl}
\usepackage{xcolor}
\usepackage{nicefrac}
\usepackage{booktabs}
\usepackage{fnpct}
\usepackage{amsfonts}
\usepackage[T1]{fontenc}
\usepackage{bold-extra}
\usepackage{amsmath}
\usepackage{amssymb}
\usepackage{bm}
\usepackage{graphicx}
\usepackage{mathtools}
\usepackage{microtype}
\usepackage{multirow}
\usepackage{multicol}
\usepackage{xpatch}
\usepackage{latexsym,comment}
\usepackage[normalem]{ulem}

\newcommand*{\missingreference}{{\Huge \colorbox{red}{?reference?}}}
\newcommand*{\missingcitation}{{\Huge \colorbox{red}{?citation?}}}

\makeatletter
\xpatchcmd{\@setref}{\bfseries}{\missingreference}{}{}
\def\@citex[#1]#2{\leavevmode
    \let\@citea\@empty
    \@cite{\@for\@citeb:=#2\do
        {\@citea\def\@citea{,\penalty\@m\ }%
            \edef\@citeb{\expandafter\@firstofone\@citeb\@empty}%
            \if@filesw\immediate\write\@auxout{\string\citation{\@citeb}}\fi
            \@ifundefined{b@\@citeb}{\hbox{\reset@font\missingcitation}%
                \G@refundefinedtrue
                \@latex@warning
                {Citation `\@citeb' on page \thepage \space undefined}}%
            {\@cite@ofmt{\csname b@\@citeb\endcsname}}}}{#1}}
\makeatother

\newcommand{\gem}[1]{\mbox{\textsc{gem}}}
\newcommand{\abr}[1]{\textsc{#1}}

\renewenvironment{quote}
{\list{}{\rightmargin\leftmargin}%
    \item\relax\small\ignorespaces}
{\unskip\unskip\endlist}

\newcommand{\hidetext}[1]{}
\newcommand{\ignore}[1]{}

\ifcomment
    \newcommand{\pinaforecomment}[3]{\colorbox{#1}{\parbox{.8\linewidth}{#2: #3}}}

    \newcommand{\prtodo}[1]{\pinaforecomment{lightblue}{pr}{#1}}
    \newcommand{\prtodoi}[1]{\pinaforecomment{lightblue}{pr}{#1}}
\else
    \newcommand{\pinaforecomment}[3]{}
    \newcommand{\prtodo}[1]{}
    \newcommand{\prtodoi}[1]{}
\fi

\newcommand{\jbgcomment}[1]{\pinaforecomment{red}{JBG}{#1}}

\newcommand{\wwcomment}[1]{\pinaforecomment{yellow}{Joy}{#1}}

\newcommand{\yzcomment}[1]{\pinaforecomment{green}{yanze}{#1}}

\newcommand{\smallurl}[1]{ \begin{tiny}\url{#1}\end{tiny}}

\definecolor{lightblue}{HTML}{3cc7ea}
\definecolor{CUgold}{HTML}{CFB87C}
\definecolor{grey}{rgb}{0.95,0.95,0.95}
\definecolor{ceil}{rgb}{0.57, 0.63, 0.81}
\definecolor{UMDred}{HTML}{ed1c24}
\definecolor{UMDyellow}{HTML}{ffc20e}

\begin{document}
\maketitle
\begin{abstract}
    % abstract

The boardgame Diplomacy is a challenging setting for
communicative and cooperative artificial intelligence.
The most prominent communicative Diplomacy \abr{ai}, Cicero, has excellent
strategic abilities, exceeding human players.
However, the best Diplomacy players master communication, not
just tactics, which is why the game has received attention as an \abr{ai}
challenge.
This work seeks to understand the degree to which Cicero succeeds at
communication.
First, we annotate in-game communication with abstract meaning
representation to separate in-game tactics from general language.
Second, we run two dozen games with humans and Cicero, totaling over 200 human-player hours of competition. 
While \abr{ai} can consistently outplay human players, \abr{ai}--Human
communication is still limited because of \abr{ai}'s difficulty with
deception and persuasion.
%current \abr{ai}--\abr{ai} and \abr{ai}-Human  communication limitations remain.
%RESOLVED \jbgcomment{Would be good to say in the abstract what the limitations are}
%
This shows that Cicero relies on strategy and has not
yet reached the full promise of communicative and cooperative \abr{ai}.

%
%Our simulations demonstrate that the success of one of these AIs, Cicero, stems primarily from its tactical moves.
%

%
%We create two new corpora of game annotations to better study the delta between game strategy and negotiation.

\end{abstract}

\jbgcomment{Add authors now!}

\section{Diplomacy Requires Communication}

In a landmark paper, \citet{meta2022human} introduce \cicero{}, an \abr{ai}
that plays the game \textit{Diplomacy}.
The Washington Post claims ``the model is adept at negotiation and
trickery''~\cite{washpost}, Forbes asserts ``\cicero{} was able to pass
as a human player''~\citep{forbes}, and even the scientific publication's editor
states \abr{ai} ``mastered
\textit{Diplomacy}''~\citep{meta2022human}.
This work tests those popular perceptions to 
rigorously evaluate the communicative and strategic capabilities of
\cicero{}.
Our observations lead to insights about the current state of
cooperation and communication in \abr{ai}, highlighting its deceptive
and persuasive characteristics.  \jbgcomment{This is a good sentiment,
  but would be better to be more direct / obvious with this, forward
  point if there isn't enough room.}  \wwcomment{added highlight}
%beginning in Section~\ref{sec:background}.
%and its place in the literature looking at communication in the game of 

\jbgcomment{
  Why was the ``AI masters Diplomacy'' in the Science journal article removed?

  Please go to: https://www.science.org/doi/10.1126/science.ade9097

  The editor explicitly added the line ``AI masters Diplomacy''.  This
  is far more trustworthy than popular press accounts.  }
\wwcomment{since AI masters Diplomacy is added, this one is resolved.}

%DENIS REMOVED THIS.  THE ARTICLE is Nature and it doesn't overclaim anything (mastering claim is about Stratego),  ~\citep{nature_diplomacy}.

%
%
%\wwcomment{I found this one from nature but it discussed Deepmind then Meta's \cicero{}  \jbgcomment{sorry, it's Science, not nature}}
%
While \cicero{} plays strategically and with a verisimilitude of human
communication, the evaluation in \citet{meta2022human} focuses only on
if \cicero{} wins games.
As the name implies, \textit{Diplomacy} is revered by its devotees as
a game of nuanced negotiation \citep{kraus1995designing}, convincing
persuasion, and judicious betrayal.
We argue that \emph{mastering} \textit{Diplomacy} requires these
communicative skills (Section~\ref{sec:background}).
Measuring persuasion, deception, and cooperation are open problems
with no clear solution.  A boardgame constrains the world of actions
to make these measurements feasible.
%However, at the time of publication \cicero{} lacked the tools to measure effectively if it had mastered these aspects of the game.  
%
One contribution of this paper is to build measurements of these
communicative skills and to evaluate the true state of \abr{ai} play
in \textit{Diplomacy}.

However, a technical challenge to identifying persuasion and deception is
mapping from communication to in-game actions (e.g., verifying that I follow through on a promise of helping you):
persuasion uses words to convince someone to do something,
and deception is saying something to alter another's belief~\cite{chrisholm1977intent}.
%RESOLVED \jbgcomment{can we add theory of deception cites?}
%
In both cases, we map and contrast communication to agents' actions.
%RESOLVED \jbgcomment{Inconsistent section references.  If you want to switch up, use a macro.}

To enable this mapping, we annotate messages from human games with
abstract meaning
representation~\citep[\abr{amr},][]{banarescu-etal-2013-abstract},
which we use to train a grounded system to infer the goals of
Diplomacy players (Section~\ref{sec:amr}).
After validating we can extract communicative intents, we use these
representations to identify persuasion and deception
(Section~\ref{sec:detection}).
We then remove some of \cicero{}'s communicative ability: 
this does not impair its ability to win games.

We then test \cicero{}'s skills in games against humans
(Section~\ref{sec:human_comp_games}) while asking players to annotate if
they think other players are an \abr{ai} and if a message is a lie.
Confirming earlier work, \cicero{} wins nearly every game, including
against top players.
However, our new annotations provide a counterargument to the
prevailing view that \cicero{} has mastered the communicative aspect of the game that is the priority of the \abr{nlp} community.
\cicero{} plays ``differently''; humans can reliably identify
\cicero{} and it is less deceptive and persuasive to human players.
Communication from \cicero{} is more transactional, relying on its
optimal strategy rather than the alliance building which is the hallmark of top human players.
\cicero{} has yet to prove effective in the communication skills which are crucial to achieving goals in strategic games and real life.

In Section~\ref{sec:conc}, we discuss what it would take for a computer to truly master
\textit{Diplomacy} and how  \textit{Diplomacy}'s intrinsic persuasion and deception can
improve computers' ability to not just talk like a human but to realistically tie words to actions.

%RESOLVED \jbgcomment{Replace this last sentence with a clearer one-sentence summary of what still needs to be done}

%RESOLVED \jbgcomment{This shouldn't be a murder mystery.  What do we think?  Did \cicero{} also show those skills?  Are they on par with humans?}

% \begin{figure}[t]
%     \centering
%     \includegraphics[width=0.4\textwidth]{figures/dialogexample.pdf}
%     \caption{Unlike previous human-\abr{AI} games, we focus on how \cicero{} uses communication. We add a feature to the Diplomacy interface so a human player can annotate when they see messages as deceptive. In this example, \cicero{}, as France, tells a human player, Italy, that they do not intend to stab them. This is annotated as a deceptive message by the human player. 
%     }
%     \jbgcomment{Should this be our first figure?  I worry this is not the best place to begin ``the story'' of the paper.  This seems more useful later when we talk about deception annotation.  If it does belong earlier, we need to make it clear why this figure is here: unlike previous human vs. \abr{ai} games, we \dots}
%     \wwcomment{Edited. (will change to width=0.4 later since comments make it too big to be in message body)}
%     \label{fig:eg}
% \end{figure}

%RESOLVED \jbgcomment{This figure could be one column (lots of whitespace).
%RESOLVED  Caption also needs to have a takeaway: why are we annotating
%RESOLVED  deception, what does this say about \cicero{}?  How does it fit into
%RESOLVED  the overall framework of the paper?}

\section{\textit{Diplomacy} as Human-AI Communication Test Bed}
\label{sec:background}

%RESOLVED \jbgcomment{Can we have a more elegant section title?}

%RESOLVED \jbgcomment{I don't think we need to actually use the term ``press''}

\textit{Diplomacy} is a strategic board game that combines negotiation
and strategy, where players take on the roles of various European
powers (nations) on the eve of World War~I.
The essence of the game lies in forming and betraying alliances to
control territories, requiring adept diplomacy (hence the name of the game) and strategic planning.
%
% This game has garnered significant interest in research for its reflection of a zero-sum game, characterized by deterministic states and actions. 
%
% 
% Diplomacy players have different play styles.
%
Some players focus on aggressive tactical decisions, while others
focus on making alliances, communicating, and collaborating with
others for better outcomes~\cite{pulsiphergames}.
The goal of the game is to capture territory, board regions called
{\bf supply centers}: once you capture enough of these supply centers, you win the game.
%\jbgcomment{Can we cite this?}

The charm and challenge of \textit{Diplomacy} messages is that players
are free to talk about anything, either strategy-related or not.
Most messages in Diplomacy are about (1)~past,
current, and future turns, (2)~alliance negotiations, and (3)~acquiring
and sharing third-party information such as, ``\textit{Did Turkey talk
  to you?}'', ``\textit{I can't believe you attacked Russia in SEV}.''
A few messages are~(4) non-Diplomacy conversation, e.g. `\textit{How are you
  today?}''

\jbgcomment{``general'' seems too broad.  This isn't a clear ontology,
  but perhaps we should split meta commentary from extra-game
  conversations}
\wwcomment{I add more cases but not sure what I can do for meta-commentary here, or maybe I could add in appendix}

\begin{figure*}[t]
    \centering
    \includegraphics[width=\textwidth]{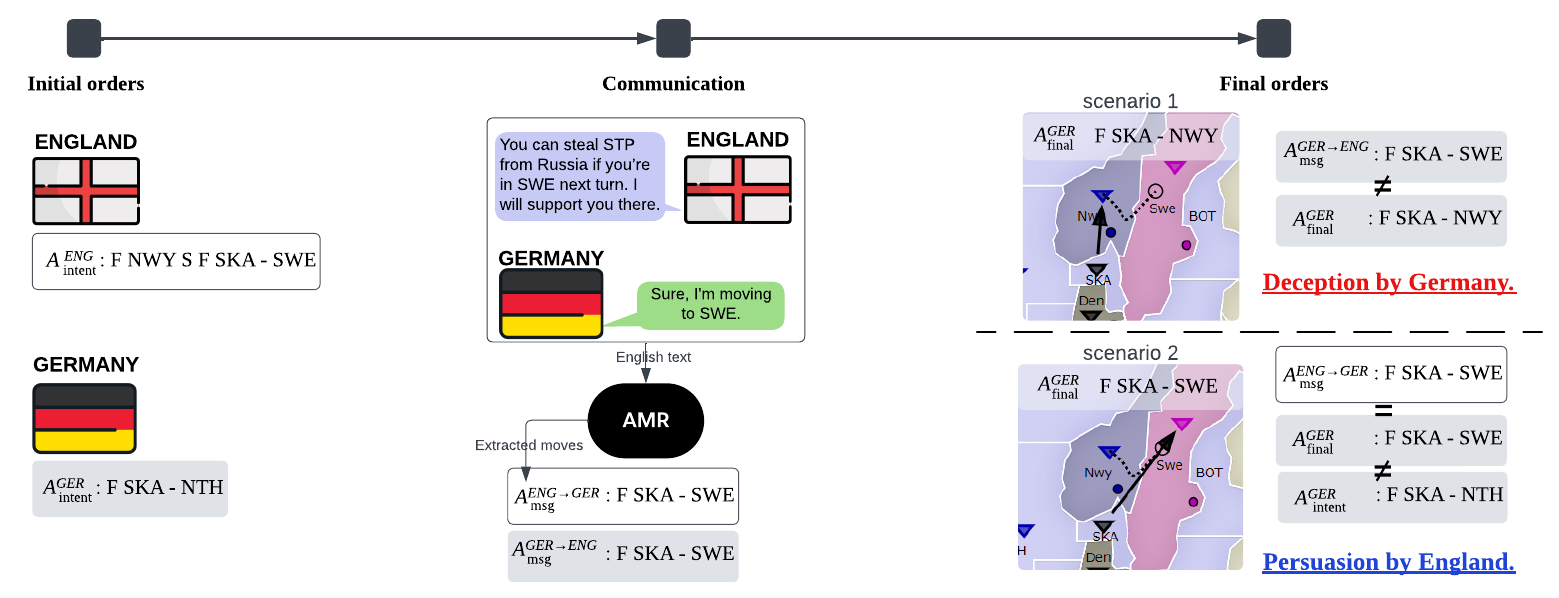}
    \caption{Our goal is to detect when players use persuasion and deception and compare human players to \cicero{}. First, we retrieve initial orders (left), then extract
      moves from natural language communication (middle) through \abr{amr}
      \ (Section~\ref{sec:amr}), and
      later detect deception and persuasion
      (Section~\ref{sec:detection}) conditioning initial intents and
      final orders (right). We show two possibilities: (top)
      Germany breaks its commitment to England by moving to Norway
      instead of Sweden, and (bottom) England successfully persuades
      Germany if Germany moves its unit to Sweden as England suggests
      and this move is not in Germany's initial orders.}
    \label{fig:dec_per_example}
    \jbgcomment{We need to make it clearer why this figure is here:
      Our goal is to detect when players successfully use persuasion
      and diplomacy and compare humans to \cicero{}. }
      \wwcomment{Edited!}
\end{figure*}

Communication in \textit{Diplomacy} is attractive to academic
researchers because it can be linguistically and cognitively complex,
but is grounded in a constrained world with well-defined states and
dynamics.\footnote{\textit{Diplomacy} without communication, sometimes
called ``gunboat'' has been studied with rule-based, \abr{rl}, and
other approaches (as we detail in more depth in
Appendix~\ref{sec:gunboat_background}).  We limit gunboat discussion here
as our focus is on the \emph{communicative} aspect of the game. }
%
% \jbgcomment{add real cites here!}
The Diplomacy \abr{ai} Development
Environment~\cite[\abr{daide}]{DAIDE} is a structured syntax for
\abr{ai} agents to play Diplomacy: create alliances, suggest moves,
etc.
Several agents have used this stripped-down communicative environment:
Albert~\cite{Albert}, SillyNegoBot~\cite{polberg2011developing},
DipBlue~\cite{ferreira2015strategic}, \textit{inter alia}.
Unlike these agents, \cicero{}~\cite{meta2022human} uses a large
language model to enable free-form \emph{English} communication,
enabling ``normal'' play with humans.
\cicero{} excelled in an online Diplomacy league, scoring over twice the
average of human participants and ranking in the top 10\% among those
who played multiple games.
In the original work, it is unclear if \cicero{}'s success is due to its use of natural language or its strategic model.
In our Human--\cicero{} studies, we ask that participants annotate messages they perceive as deceptive, allowing us to more carefully study the communicative aspects of the game (more details are in Section~\ref{sec:human_comp_games}).
% (Figure~\ref{fig:eg}, more details in Section~\ref{sec:human_comp_games}).
%
% Figure~\ref{fig:eg} shows one example of human annotations that a human player (Italy) annotates \cicero{}'s message (France) as deceptive.
\jbgcomment{I'm not sure what this is trying to show: that it says
  false things because of LLM hallucinations?  That we're measuring
  it?  Let's be clearer about why we're giving this example.}
\wwcomment{I edited the caption and what we trying to show here}
\subsection{\textit{Diplomacy} without Communication is not \textit{Diplomacy}}
\label{sec:communication}
Having discussed the basics of \textit{Diplomacy}, we now
turn to what makes the game unique.
Because the game is relatively balanced between seven players at the
start, players need to form alliances if they hope to gain an advantage.
However, these alliances should be mutually beneficial; from a
player's perspective, they need to advocate for cooperation that 
benefits themselves.
This requires effective \emph{persuasion}~\cite{cialdini-00}: making
appeals to scarcity, reciprocity~\cite{kramar2022negotiation}, unity, or shared norms.
This is a communicative task which involves social and emotional skill: picking the right moves and convincing other players to help them.

However, the ultimate goal of \textit{Diplomacy} is for \emph{individual} players to
win the game.
This means that alliances will fall apart, leading to \emph{deception}~\cite{peskov2020takes}
as part of a betrayal~\cite{Niculae:Kumar:Boyd-Graber:Danescu-Niculescu-Mizil-2015}.
Because a player might benefit from a victim thinking that they are
working together, a betrayer often sets up the
tactical conditions for a betrayal while obfuscating their goals through cleverly composed deceptive messages (even if not outright lies).

Because deception and betrayal are communicative acts both necessary for mastering and enjoying
\textit{Diplomacy} and grounded in the
state of the game, the next sections develop tools to detect when
they happen.
This will allow us to measure whether \abr{ai} agents like \cicero{}
have mastered both tactics and communication.

\jbgcomment{This section seems less ``tight'' than some of the others,
  it could proably shortened given the some of the other points are
  raised elsewhere.}

\section{Grounding Messages Into  Intentions}

\label{sec:amr}
% \jbgcomment{I made up an example, would be nice if we could find a real one\dots}
Consider this in-game statement made by England to Germany about a
specific move-set (glosses added to locations):
\begin{quote}
    % I'm going to stand pat in MAO, you should totally nab Spain while you can
    % I manage to hold NWY, you should grab Sweden if that's more to your liking.
    You can steal STP [St. Petersburg] from Russia if you’re in SWE [Sweden] next turn. I will support you there.
\end{quote}
We want to be able to tell if the speaker is lying (e.g., they're
going to do something else instead of what they claim they're going to
do) and if the speaker has convinced the recipient to alter their
actions.
This is necessary to measure how effectively \cicero{} communicates in the game.

While we know the intended \emph{actions} of players when they submit their
moves, we need to see how those moves match up to their
\emph{communications} in the discussion period before they submit
moves (Figure~\ref{fig:dec_per_example}). 
We use \abr{amr} to build a
machine-readable representation of \textit{the intent of actions} in their communications. 
We are not starting from scratch: \abr{daide} (Section~\ref{sec:background}) provides a set of predicates (ally, move, etc.) critical to Diplomacy communications.
We thus focus on annotating these predicates that encode actions, allowing us to understand the \emph{communicative intent} of messages, where speakers could say they will do something \textbf{and} follow through, or say they will do something and \textbf{not} follow through.
% 
% Our goal of detecting deception and persuasion requires grounding what people say into the same space as what they do.  
%
% In the case of \textit{Diplomacy}, these are communications between players and moves on the board. 
%
% With a well-known common limitation that humans do not always convey information about their intentions, the press in Diplomacy is not different; it is also in the form of delivering intentions of moves with various latent strategies; a player may variously perceive, persuade, or even deceive. 
%
% \jbgcomment{I think we need to ground this ourselves a little bit more: have a motivating example (perhaps as a figure) of what is persuasion and what is deception}
% RESOLVED \jbgcomment{I think Figure 1 / 2 miss a big part of our pipeline which is checking the extracted AMR / DAIDE whatever from the English communications and then comparing them to the moves and the user's pre-move plans.  I think that would make for a stronger picture.

%   (e.g., something like slide 15 from %https://docs.google.com/presentation/d/1WaHLRZMqMIB5fGje1J2T0adCJy7Ueh0WUZeNqu1FNlI/edit#slide=id.g2523761b7b9_0_10)

%   This figure should have: raw English, extracted logical form, the pre-season intents, final moves, comparison of the two.
% }
% \wwcomment{I updated Figure 2 and ref it in section 3 and 4}
%
% This section describes our annotation process to build a dataset to
% fine-tune \abr{amr} parsers grounded in the game of \emph{Diplomacy},
% which we then use to extract \textit{actions} and detect persuasion and deception in
% Section~\ref{sec:detection}.
%
Because not all information needed for annotation is in the
raw message text, we further show human annotators who wrote the text (e.g., France,
Germany), seasons (e.g., Spring 1901), and the current game state.
This information is necessary to annotate \textit{``You can steal
  STP from Russia if you are in SWE this turn. I will support you there''} in the earlier example so that the annotators can assume what unit would support and what unit would move into Sweden.
In this case, England's fleet in Norway supports a German fleet in Skagerrak to move into Sweden.
\jbgcomment{Add something like: In this case, a Russian fleet in\dots supported by a German army in \dots}
\wwcomment{Added.}
  
%Overall, the annotation facilitates the persuasion analysis in
% Section~\ref{sec:persuasion-detection}, which answers if England was
% ultimately successful in persuading Germany.
\wwcomment{I want to focus on implicit action in communication, so I removed the connection to deception/persuasion here (we already have a connection in section 3.2). We may remove the last paragraph here if this additional info for annotations isn't important for AMR intro.}
% This approach differs from previous efforts in \abr{amr} annotation, such as those by \citet{banarescu-etal-2013-abstract}, which aimed to annotate the entire sentences without filtering for content relevance.

%RESOLVED \jbgcomment{make then parentheticals consistent.  Should be e.g.,France/Germany for speaker / recipient, right?}

% 
% We further instruct
% annotators to only annotate for \textit{communicative intent}, that
% is, information that is directly relevant to the game. 

% This departs from some earlier \abr{amr} annotation
% efforts~\cite{banarescu-etal-2013-abstract} which attempted
% full-sentence AMR annotation regardless of content.

\jbgcomment{This sentence doesn't make sense to me without more context}
\yzcomment{Restructured the above sentences to make context more reasonable.}

\subsection{Annotation}
\label{sec:amr_annotation}

Like any specialized domain, \textit{Diplomacy} has its unique
vocabulary.
%
% We extend the \abr{amr} vocabulary not just with abbreviations like ``NWY'' (Norway) but also with verbs like ``threaten'' or ``demilitarize'' (set up a demilitarized zone) and to cover gaining/holding/losing provinces, especially supply centers (``SC'') in the example (Full extended vocabulary in Appendix~\ref{sec:amr_annotation_appendix}).
Taking the above statement as an example, we extend the \abr{amr}
vocabulary to include not only abbreviations, such as ``SWE'' for
Sweden, but also verbs like ``threaten'' and ``demilitarize'' (to set
up a demilitarized zone), as well as to describe actions like gaining,
holding, or losing provinces, especially supply centers (``SC''),
which are equivalent to points and integral to winning the game.
\jbgcomment{It seems like we use supply centers elsewhere (for the analysis), so we might want to be a little more deliberate about how we introduce it.  E.g., make it clearer that this is ``how you win the game'' in the Diplomacy section and then make something we can backpoint to when necessary.}
\wwcomment{I see you nicely introduced supply centers in Diplomacy section, I assume this is resolved.}
In contrast to standard \abr{amr} annotation, where every sentence is fully annotated, Diplomacy \abr{amr} annotation sometimes involves only partial or no annotation for certain utterances, depending on their relevance to gameplay strategies like forming alliances or making moves, exemplified by \abr{amr} concepts such as \amr{ally-01}, \amr{move-01}, and \amr{attack-01} (the full extended vocabulary introduction in
Appendix~\ref{sec:amr_annotation_appendix}).

% \jbgcomment{
% Don't need all of this, move to appendix with full dictionary:

% Building on general AMR annotation guidelines, we established additional Diplomacy-specific AMR annotation guidelines, including what and how to annotate. Unlike general AMR annotations, where all sentences are fully annotated, in Diplomacy AMR annotations, some utterances are only partially (or even not at all) annotated, based on the degree of usefulness for Diplomacy.
% % % https://www.isi.edu/~ulf/amr/lib/amr-dict-diplomacy.html

% We analyzed player messages for additional concepts of high Diplomacy communication value, and extended the Diplomacy AMR vocabulary by including concepts crucial for conveying alliances, betrayals, and strategic maneuvers, which allows annotators to easily mark sentences. We also extended AMR guidelines to cover gaining/holding/losing provinces, especially support centers. 

% The general AMR Editor includes a Checker that performs a battery of tests to ensure well-formed and consistent AMRs. We extended the Checker for Diplomacy AMRs by listing AMR concepts (e.g. \amr{betray-01}), their related English terms (e.g. \ betray, stab, traitor, treason), annotation examples, any corresponding code for further usage, and notes.
% % , e.g.\ to ensure that for a \amr{build-01}, the location is an argument of \amr{build-01} itself, rather than an argument of the army or fleet being built.

% }

%\jbgcomment{can we get agreement numbers?}

In a preliminary annotation phase, we have Diplomacy experts annotate sentences from \citet{peskov2020takes} to train human annotators and refine the Diplomacy Appendix of the \abr{amr} Annotation Dictionary.
%
%Our lead Diplomacy \abr{amr} annotators 
We annotate 8,878 utterances
(ranging from a word to several sentences).
4,412 of those utterances are annotated as empty \abr{amr}s (e.g. for
``Lemme think about your idea'') indicating no in-game move intent.
598 of the annotated \abr{amr}s contain full information
extracted from messages of Diplomacy games. The remaining 3,868 \abr{amr}s
contain 3,306 utterances with underspecified information such as units
with missing type, location (Figure~\ref{fig:amr_loc}), and nationality (Figure~\ref{fig:amr_country}), as well as 562 agreements with a
missing object.
Many utterances contain underspecified information, as Diplomacy players often communicate with messages that lack specific details (which are implicit and can be inferred from the game state).
\wwcomment{adding this sentence to clarify questions from two reviewers}
The annotated \abr{amr} corpus is further used for
training our English-to-\abr{amr} parser to extract communicative intent information from
utterances.

\jbgcomment{It might be good to have a little more detail in the appendix and point to that here.}
\wwcomment{point to figure A1 and A2 where paring English to AMR with underspecified location and country}

\subsection{Training a Parser to Detect Intentions}
\label{sec:amr_parser}

We use a sequence-to-sequence
model~\cite{amrlib2023}
fine-tuned on the \abr{amr} 3.0 dataset~\cite{LDC2020} as the baseline model to detect
communicative intent in new conversations.
% \jbgcomment{Can we upgrade the footnote to a real cite?}
% \fgcomment{There is no technical paper on the models other than that link. should we just use a specific entry type/cite seq2seq?}
% \yzcomment{I will cite the whole repository later, I think github has this support.}
%
Following other \abr{amr} work, we report parsing accuracy via the
widely-used \abr{smatch} score~\cite{cai-knight-2013-smatch}.
We divide the annotated corpus into 5,054 training, 1,415 validation
and 2,354 test sentences, where each sentence is in the train /
validation / test folds, split by game~\citep{peskov2020takes}.
% which
% ensures we generalize across speakers and phenomena specific to a
% game. In other words, the validation and test folds are from discrete
% games.

% \jbgcomment{Describe pipeline for detecting AMR from text}

% While stylistic aspects play an indispensable role in sustaining engagement and interest among participants, factual information is more vital for informed decision-making. Detecting deception and persuasion in communications requires checking the relationship between message information and initial/final moves of a particular power. To address the nuanced challenge of distinguishing meaningful and informative content from stylistic dialogues in \textit{Diplomacy}, our focus herein is on developing a sophisticated pipeline for information extraction using AMR from text. We propose an innovative approach that incorporates valuable world knowledge into our system, thereby augmenting its capability to parse and interpret the multifaceted nature of dialogues. Semantic Parsing for natural languages to AMR is well-established. For \textit{AMR Baseline Semantic Extraction}, we utilize pretrained Sequence-to-Sequence model from the Huggingface transformers library. This model has been fine-tuned using the AMR 3.0 (LDC2020T02) dataset \cite{} for parsing in amrlib. We employ SMATCH  score to evaluate AMR accuracy, which is the most commonly used metric for evaluating AMR parsers. 
%

We improve the model, starting from a baseline version without
fine-tuning with \abr{smatch} of 22.8.
Our domain-tuned model using Diplomacy-\abr{amr} improves \abr{smatch}
by 39.1, to 61.9.
Adding data augmentation into the model (e.g., knowing the sender of a
message is England and the recipient is Germany) improves \abr{smatch} to 64.6.
Adding separate encodings for this information further improves
\abr{smatch} by 0.8 (65.4).
Additionally, we apply data processing to replace (1) pronouns with country names and (2) provinces in abbreviations with full names, which increases \abr{smatch} to 66.6 (More parser details in
Appendix~\ref{sec:amr_parser_appendix}).
This parser enables us to evaluate the role that communication has in
\cicero{}'s capabilities.

\jbgcomment{if we have new AMR improvements, this should be updated with those}
\wwcomment{updated}

\subsection{Does \cicero{} need to Talk to Itself to Win?}
\label{sec:Comp_comp_games}

\jbgcomment{I think we could beef up this argument a little bit.
  Perhaps we could talk about an easier question: does communication
  matter?  This was answered in Science paper.  But we're interested in a different question: what communication matters, and does \cicero{} do it.

The current setup of the section sounds more like we care about
testing whether communication matters at all, which is not what we're
doing.}
\yzcomment{update subsection title and first, second paragraphs to focus on Jordan's suggestions. Thank you!}

% It is natural to question how \cicero{} use its communicative ability to help win a game. We want to measure what communication matters in Diplomacy game and does \cicero{} really make use of it. 

% the importance of communication in a game
% that can be played without any communication, which is known as
% `gunboat'---they cannot send messages and cannot read messages (Appendix~\ref{sec:ext_related}).

% \jbgcomment{we have an appendix section on that, forward point to that}
%
% We answer this question by running a series of games where all the
% players are variants of \cicero{}, with modifications to their
% communication abilities, from `gunboat'---they cannot send messages and cannot read messages (Appendix~\ref{sec:ext_related}) to Natural Language level.

The first question is whether communication matters in games with other \cicero{} agents (deferring the question of competition with humans to later sections).
We have \cicero{} variants with different levels of communication abilities---ranging from ``gunboat'' without any messages (Appendix~\ref{sec:gunboat_background}) to full Natural Language capabilities---play each other and evaluate the results.

%
% With the \abr{amr} parser in hand, we are able to degrade \cicero{} so that it is only able to communicate with
% transactional speech that aligns with a subset of \abr{amr} consistent
% with the communicative protocols in
% Appendix~\ref{sec:amr_annotation_appendix}.
% With the \abr{amr} parser in hand, we are able to degrade \cicero{} so that it can only use speech that corresponds with a subset of \abr{amr} and is in line with established communicative protocols (Appendix~\ref{sec:amr_annotation_appendix}).
%
% In other words, we run the parser over utterances, transmit anything
% that we can parse, and throw everything else out.
% 
For the seven \cicero{} variants in each game, we randomly select three to have communicative
ability; the remaining four play communication-less ``gunboat.''
The selected three communicative powers have the same
\textit{communication level}. 
We define a set of communication
levels, from more communicative to less
communicative:
% \jbgcomment{Some thoughts on this section:
% \begin{enumerate*}
%     \item Call the levels something interpretable
%     \item Figure that fits in one column is better than giant table
%     \item We need to be clearer about why we're doing this.  Something like: by restricting \cicero{}'s communications to be purely transactional, we remove the possibility of rhetorical strategies to influence other players
% \end{enumerate*}
% }
%RESOLVED \jbgcomment{Cut passive voice}
\begin{itemize*}    
    \item \abr{Natural Language}: the \cicero{} agent of \citet{meta2022human} with full natural language
    \item \abr{amr}: only messages about game actions (i.e. those that are parsed by \abr{amr}) go through, allowing the agents to coordinate game actions (Appendix~\ref{sec:amr_annotation_appendix})
    %they communicate using extracted complete information, attempting to isolate communicative intent % but is inherently noisy due to the processing pipeline.
    \item \abr{Random}: a random message from a corpus of previous Diplomacy games is sent,\footnote{We match the assigned power of the sender and receiver and the year, which makes the message slightly more convincing (but unlikely to be consistent with the game state).} mimicking form without content. %whatsoever
\end{itemize*}
%
% Limiting \cicero{} to this minimal communication does not significantly impede its ability to cooperate and communicate with itself. However, it's possible that \cicero{}'s natural language skills might not be as effective on a fellow AI. 
% Limiting \cicero{} to this minimal communication does not significantly impede its ability of building the strategy. It's possible that \cicero{}'s natural language skills might not be such effective on a fellow AI.
% Also, it is possible that \cicero{}'s natural
% language skills are not such effective as a key to winning.
%RESOLVED \jbgcomment{Have we defined ``power''?}
\cicero{} plays 180 games with itself; 60 games for each communication
level.
In each game, we stop after 14 movement turns with 10 minutes
of communication for each turn.
%
% The power distribution of three randomly picked communicative powers
% is balanced, with each power being selected 24--28 times across the
% three different communication levels.
%
We randomly select which power the agents are assigned to, so  power distribution is balanced.

\begin{figure}[t]
    \centering
    \includegraphics[width=0.45\textwidth]{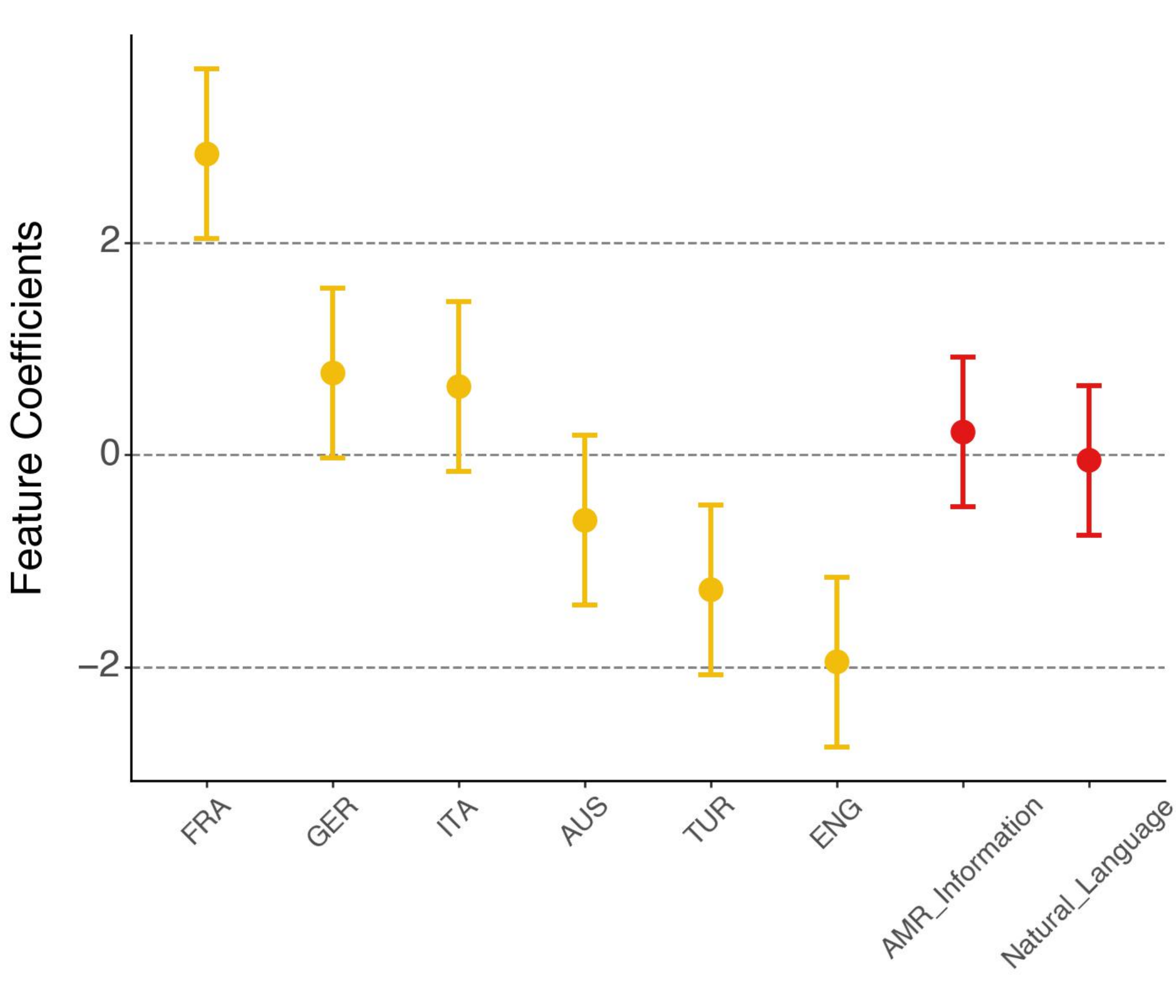}
    \caption{Power assignment is strongly predictive of \cicero{} performance as measured by supply center gains. Coefficients (with 95\% confidence intervals) from a linear regression with \abr{Random} Message/Russia as a baseline, show that the effect of choosing the effect of changing language systems is trivial compared to changing powers.}
\label{fig:Feature_coef}
\end{figure}

We measure performance by the number of supply centers (and thus how well the agent played the game, Appendix~\ref{sec:amr_CCgames_appendix}).
Consistent with our hypothesis that performance is driven by tactics, the gains \cicero{} gets from communication is substantially smaller than the gains from playing a stronger power (Figure~\ref{fig:Feature_coef}):
Playing as France (\abr{fra}) yields an expected 2.8 additional supply centers (2.0--3.6 95\% interval)  compared to the median power Russia (\abr{rus}). In contrast, the best language condition \abr{amr} only yielded an expected 0.2 additional supply centers (-0.5--0.9 95\% interval). In other words, the effect of choosing the best power over the median power is 14 times larger than the best communication strategy. \jbgcomment{Could we use different typesetting for the conditions?}\wwcomment{done.}
This is consistent with prior findings~\citep{sharp1978game} that France is the easiest power to play and our other findings that \cicero{}'s  communicative ability plays no clear role in its win rate.
% \jbgcomment{citation?} \wwcomment{can find only one book though many online blogs (reddit)}
%
% The 95\% confidence intervals of the coefficient for France ranged from 7.05 to 8.08 and for Germany from 4.98 to 6.01, indicating a high level of precision in these estimates. 
%
% In contrast, communication strategies such as  
%

To better understand what \cicero{} is using communication for, we build on our \abr{amr} representations to capture intent in the next section.

% Figure~\ref{fig:Feature_coef} shows the feature Coefficients in the regression model. From these results it is clear that \cicero{} doesn't really rely on its communicative skills compared with its strong decision-making strategies.

% To better understand \cicero{}'s communicative abilities with humans, we first need to define what constitutes persuasion and deception, building on our \abr{amr} dialog representation.
% Results are shown in Figure \ref{fig:Feature_coef}.

\section{Promises Made, Promises Kept, and Finding Dirty Lies}
\label{sec:detection}
% highlevel intro
% how do we do for persuasion and how do we do with deception

% Deception and persuasion are complex facets of human interaction and essential mechanisms within the game of Diplomacy for strategic success and integral to its gameplay intricacies.
% %
% These elements often manifest indirectly through alliance dynamics, yet capturing them poses a challenge due to their unclear pattern.
% %
% Despite these limitations, lower-level detection, like analyzing player moves, offers a comprehensive method that can reveal high-level deception and persuasion.
% %
% Betraying alliances or misleading future strategies reflect the core of deception, while persuasion is visible in moves aimed at influencing others' actions. 
% %
% Thus, move-level detection emerges as a potential approach for capturing strategic deception and persuasion in Diplomacy. 
%
%resituate again for deception and persuasion
%connect to cicero paper a bit - 
% 
In the example from the previous section (Figure~\ref{fig:dec_per_example}), England
says that it would support Germany's move into Sweden from the
Skagerrak Sea, while Germany agrees with the proposal from England.
What does it mean for these to be deceptive or persuasive?

The geopolitical definition of deception is to manipulate
adversaries' perceptions to gain strategic
advantage~\citep{deception1982}, e.g., Germany alters England's belief
so that England would do an action that benefits Germany (i.e. Germany
has a better chance of ending up with higher score).
However, evaluating deception is challenging because it requires
estimating the differences in England's beliefs \textit{before} and \textit{after} Germany deceives
England.
Therefore, we break down a broader, amorphous concept into easier-to-handle concepts and leave broader deception to future work.
The first subconcept is {\bf breaking of a commitment}~\citep {kramar2022negotiation}: someone saying they will do something and not following through.
In the example, if Germany commits to moving to Sweden but later attacks England in Norway, this will be detected as a \textit{broken commitment}.

\jbgcomment{We need to make sure that the language between what we
  asked people to annotate (and those instructions) are consistent.
  E.g., the figure has ``is this message deceptive'', while this is
  talking more about whether the message is a ``lie''.}
\wwcomment{I changed the language to be deceptive instead of lie for human annotation}

The second subconcept is {\bf lying}: human players in games with \cicero{} annotate messages their messages as either \textit{truthful} or \textit{deceptive}.
We build on~\citet{peskov2020takes}, who define deception to players thusly: ``\textit{Typically, when [someone] lies [they] say what [they] know to be false in an
attempt to deceive the listener.}''
Again, deception is broader than lying (and there are top Diplomacy players who intentionally deceive while never outright saying anything untrue), 
and our definition of deception is slightly broader than \citet{peskov2020takes}.
In our more permissive annotations, 
human players mark interactions that are \textit{broken commitments}, lies about other players, hedging about the state of alliances, and anything else that they feel is deceptive.
Despite this ontological uncertainty, for convenience, we refer to the process of humans annotating messages as ``\textit{human lie annotations}'' for consistency with \citet{peskov2020takes}.
To recap: a broken commitment can be a lie and an example of deception.\footnote{Technically, not all broken commitments are lies: it could be an honest mistake.  It's also possible that a player says that they ``typed in their order wrong, sorry!'' which is a {\bf lie} to cover up~\citep{deception1982} a {\bf broken commitment} as part of a broader deception strategy.}
Likewise, not all lies are broken commitments, but both breaking a commitment and lying are facets of deception. 
While we cannot capture all deception---because it is based on internal state---it is important to capture as much as we can to measure how important it is to playing \textit{Diplomacy} well.
% (See Figure~\ref{fig:dec_definition}).
%
% Moreover, annotating \textit{deceptive} in received messages can be seen as
% an \textit{estimated} broken commitment using human judgment.

Specifically, \cicero{}'s ability (or inability) to decieve or persuade has never been empirically
measured, so we build on the \abr{amr} parser of the previous section
to detect broken commitment and persuasion.
As we discuss in Section~\ref{sec:amr_parser}, this is not perfect,
but it has good coverage when players discuss their
intentions.
% Detecting deception and persuasion is a difficult task since humans have complex ways of communicating through texts.
%
% Though the Diplomacy game has explicit gameplay moves that can often differentiate cooperation and betrayal, players can exchange texts or press in open-ended ways to align their best interests and a belief to win at the end.
%
% We address this challenge of extracting persuasion and deception from texts and defining them simply using AMR and rule-based functions.
%
% We also specially designed a game interface for human players to annotate \textit{whether the message is a lie} for the message senders and \textit{whether the message looks like a lie} for the message recipients (Interface details in Section \ref{sec:games}). 
%
First, we parse English messages to \abr{amr} structures, for which we define
actions that the speaker intends to do, e.g. we can extract Germany’s \texttt{communicative intent} (\texttt{F SKA - SWE}) when Germany agrees with
England that they will move to Sweden  (middle, Figure~\ref{fig:dec_per_example}).
We also define orders players submit before any communication as \texttt{initial intents} and \texttt{final orders} as orders that players submit when the turn ends.

Using \texttt{initial intents}, \texttt{communicative intent}, and \texttt{final orders}, we can now define broken commitment and persuasion.
For a broken commitment, we say that Germany violates a commitment with England if Germany verbally agrees to move to Sweden but actually attacks England in Norway 
(Deception by Germany, Figure~\ref{fig:dec_per_example}).
Breaking a commitment may result when an intent changes but is not communicated. 
For example, Germany agrees with England to move to Sweden but instead moves to Denmark to defend against France without informing England.
Although this may not involve deceptive intent, we still consider it deception because it alters the listener's beliefs and affects decision-making. 
For instance, England might decide to support Germany based on their agreement.
For persuasion, England's request is considered persuasive if Germany moves to Sweden, as England suggests, instead of Germany's original plan to move to the North Sea (Persuasion by England, Figure~\ref{fig:dec_per_example}).
\jbgcomment{This was marked as resolved, but I don't think it was: Where did we introduce this?  If we did, give backpointer.  If not, needs to be introduced more elegantly}
\wwcomment{We haven't introduce, so I improved the intro here}
We describe each of these more formally in this section.
% moves from English texts using AMR (from section \ref{sec:amr}), which can be represented in a graph structure. 
% as displayed,

% \noindent {\scriptsize \texttt{(m / move-01\\[-1.5mm]
% \hspace*{4 mm} :ARG1 (a / army\\[-1.5mm]
% \hspace*{8 mm}            :mod (c / country :name (n / name :op1 "France"))\\[-1.5mm]
% \hspace*{8 mm}            :location (p / province :name (n2 / name :op1 "Paris")))\\[-1.5mm]
% \hspace*{4 mm} :ARG2 (p2 / province :name (n3 / name :op1 "Burgundy")))}}\\[1mm]
% %
% In this example, we can extract \textit{France army in Paris is moving to Burgundy}, which is equivalent to the move \texttt{A PAR - BUR} in the \citet{NEURIPS2019_84b20b1f}'s game engine.
%
% The important components in the detection are:
% \begin{itemize}
%    \item Intents
    %\item extracted moves
    %\item final orders
% \end{itemize}
%
% As we already introduced, the \texttt{extracted moves} are moves we extract from conversations using AMR.
\subsection{Broken commitment} 
\label{sec:deception-detection}

\jbgcomment{1.  Shouldn't this actually be a set membership operation
  rather than equality?  I.e., we don't expect them to say all of
  their moves, but the things that they say they'll do should be in
  there?}

\jbgcomment{I think we can also integrate this equation better into
  the sentence}
\wwcomment{updated to a set of final orders and a set of initial intents (in persuasion)}

We define broken commitment in Diplomacy when a player~$i$ commits to doing an
action~$a^{i\to j}_{\text{msg}}$ and does not do it. In other words, given a set of \texttt{final orders}~$\mathbf{A}^{i}_{\text{final}}$ from player $i$, 
if $a^{i\to j}_{\text{msg}} \notin \mathbf{A}^{i}_{\text{final}}$, then this is a broken commitment, i.e., 
\begin{equation}
    \text{BC}(a^{i\to j}_{\text{msg}},\mathbf{A}^{i}_{\text{final}}) = 
    \begin{cases}
        1,      & \text{if } a^{i\to j}_{\text{msg}} \notin \mathbf{A}^{i}_{\text{final}}\\
        0,      & \text{otherwise.}
    \end{cases}
    \label{eq:bc}
\end{equation}
Note that a player~$i$ agreeing to 
player~$j$'s \textit{proposal} to do
action~$a^{i\to j}_{\text{msg}}$ is equivalent to directly committing to \textit{doing} that
action.
\jbgcomment{Can we give an example of that with the same level of specificity?}
\wwcomment{not sure if I understand this right, but I add terminologies to it}

\subsection{Persuasion}
\label{sec:persuasion-detection}

Broken commitment is in some ways easier to detect than persuasion, as we are
only comparing a spoken intent to a final action.
Persuasion is more difficult because we must discover initial
intents, then compare them to communication \emph{and} to final moves.

Because we want to be able to measure persuasion for both humans and
for \cicero{}, we need comparable representations of \texttt{initial intents}
for both.
%
%solved \jbgcomment{Cite equation/section}
Thankfully, \cicero{}'s architecture uses a conditional language
model~\citep[Equation S2, section D.2]{meta2022human} that generates
its natural language messages given a set of moves (e.g., France
internally decides it will
do \texttt{F MAO - POR, A BUR - MAR and A MAR - PIE}) and then its
messages reflect those \emph{intents}. We directly use this set of intents from \cicero{} as \texttt{initial intents} in the persuasion detection.
For humans, we explicitly ask all players to provide their planned
moves (i.e., the same information that \cicero{} uses in its internal
representation) before the negotiation turn begins
(Section~\ref{sec:human_comp_games}).
In other words, we ask humans to directly input their intent, unlike \cicero{}, where we log its computational intent. 

%
% The \texttt{final orders} are \textit{orders} that players submitted in each phase. 
%
% Lastly, the \texttt{intents} are defined by the \textbf{first} moves players intend to play.
%

% \cicero{} has a move, e.g.,  as a control code for its language model so that it will generate such English text to converse with other players . 
%
% By tracking \cicero{}'s internal moves during experiments, initial moves from \cicero{} are the intent in each phase.
%
% \cicero{} is clear about its intent, but humans are not.
%
 %Without initial moves like \cicero{}, we replicate intent tracking by asking the participants what they would do if they had to make moves before any communication.
%
% The first moves that human players select are defined as intent.
% We keep track of all moves that human players select, so the first moves humans select are defined as initial moves.
%

% \jbgcomment{Add math in the below sentence after "move" and "intent"}

Persuasion happens when player~$i$ talks to player~$j$, suggests an
action~$a^{i\to j}_{\text{msg}}$, and then player~$j$ makes a set of
\texttt{final orders}~$\mathbf{A}^{j}_{\text{final}}$ that is different from 
their \texttt{initial intents}~$\mathbf{A}^{j}_{\text{intent}}$. In other words, player~$j$ is persuaded by player~$i$ if they commit an action suggested by  player~$i$, $a^{i\to j}_{\text{msg}} \in \mathbf{A}^{j}_{\text{final}}$ that was not player~$j$'s initial intent $a^{i\to j}_{\text{msg}} \notin \mathbf{A}^{j}_{\text{intent}}$.
We define persuasion $\text{Per}(\mathbf{A}^{j}_{\text{intent}}, a^{i\to j}_{\text{msg}}, \mathbf{A}^{j}_{\text{final}})$
% By considering a private conversation of each pair of powers, we gather a list of intents $\mathbf{A}_{\text{intent}}$, extracted moves $\mathbf{A}_{\text{msg}}$ and final orders $\mathbf{A}_{\text{final}}$ from dataset conversations.
%
% Each pair of powers can be represented as ($p_i$, $p_j$) where $i \neq j$ and $p_i,p_j \in \mathbf{P}$ and $\mathbf{P}$ is a set of seven countries in Diplomacy standard map.
%
% 
% Persuasion is grounded slightly differently than deception.
%
% If a player $p_i$ asks or suggests another player $p_j$ to do an action $A^{i\to j}_{\text{msg}}$ in the conversation, and that player $p_j$ follows $A^{j}_{\text{final}} = A^{i\to j}_{\text{msg}}$, we can consider this event as persuasion. 
%
% However, we notice that one could initialize the moves, which can be similar to a request or suggestion from the other. 
%
% In that case, this could be considered persuasion, though there is no \textit{persuasion} in conversation that makes a difference in moves.
%
% Therefore, we deliberately define a \textit{persuasion} function $Per(\cdot)$ in Diplomacy as

\jbgcomment{This is oddly formatted.  Can we move the lefthand side to
the body of the text and just have the braces here?  I think that we
also need to think about set inclusion, not strict equality here.}
\wwcomment{fixed!}
\begin{equation}
     % \text{Per}(\mathbf{A}^{j}_{\text{intent}}, a^{i\to j}_{\text{msg}}, \mathbf{A}^{j}_{\text{final}})
     = \\
    \begin{cases}
    1, & 
    \begin{aligned}[c]
    &\text{if } a^{i\to j}_{\text{msg}} \in \mathbf{A}^{j}_{\text{final}} \\
    &\text{and } a^{i\to j}_{\text{msg}} \notin \mathbf{A}^{j}_{\text{intent}},
    \end{aligned} \\
    0, & \text{otherwise.}
    \end{cases}
    \label{eq: per}
\end{equation}

\section{Comparing \cicero{} to Humans}
\label{sec:human_comp_games}
\begin{table}[t]
\centering
\begin{tabular}{lrrr}
\hline
 & \cicero{}&Human&\textbf{Total}\\ 
\hline
 Players&99&69&\textbf{168}\\
 Messages& 20270&7395&\textbf{27665}\\
\hspace{3mm} annotated as lie&-&318&\textbf{318}\\
\hspace{3mm} perceived as lie&-&1167&\textbf{1167}\\
 Intents&2632&1328&\textbf{3960}\\
\hline
\end{tabular}
\caption{Overall statistics of Diplomacy dataset that we collect across 24 Human-\cicero{} games, including (1) number of human players and number of times \cicero{} plays, (2) total messages sent by humans and \cicero{}, (3) lies annotation where humans send lies and perceived as lies (4) total initial intents from \cicero{} and humans}
\label{tab:overall_stat}
\end{table}

\jbgcomment{we seem to go back and forth between strategy and tactics
  ... which is it?}
\wwcomment{strategy is more consistent to Meta's paper}

\cicero{} has strong strategic abilities and is relatively cooperative towards other players (Section~\ref{sec:Comp_comp_games}), but it is unclear whether \cicero{} can
achieve human-level gameplay in both tactics and
communication. Having defined the aspects of communication that we
argue are important for mastering \textit{Diplomacy}, we want to
investigate communication and cooperation between \cicero{} and
humans. Specifically, we want to answer:
\begin{enumerate*}
    \item Can \cicero{} persuade humans?
    \item How deceptive is \cicero{} compared to humans? 
    \item Can \cicero{} pass as a human?
\end{enumerate*}

%RESOLVED \jbgcomment{Here is where the annotation is first introduced, but it's necessary earlier.  We need to introduce it at least a high level earlier.}
% UPDATED: lie annotation was introduced in intro

%SOLVED \jbgcomment{I think this would also be stronger if this took the form of talking about why we do something rather than just saying we do. Some of this isn't particularly relevant.  We can be honest about that and perhaps have a whole section about that in the appendix.}

We adapt the game engine created by \citet{paquette-19} and introduce
additional measures  to the interface to help us answer these
questions.  To measure if human players are persuaded, we record their moves before communication starts ($\mathbf{A}^{i}_{\text{intent}}$ in Equation~\ref{eq: per}). 
Following \citet{peskov2020takes}, humans annotate every message that they receive or send: 
they annotate each outgoing message for whether it is a lie (truth/lie/neutral options),
and they annotate each incoming message for whether they perceive it as a lie (truth/lie options).
% incoming and outgoing message for whether it is a lie (outgoing) or In a joint effort
% with the parser (Section~\ref{sec:amr_parser}) to better capture how
% humans and \cicero{} deceive differently, humans annotate all incoming and outgoing messages\footnote{The option % is binary (deceptive/not deceptive) for incoming messages and ternary  for outgoing ones.} (human lie annotation). 
%
% For convenience, we refer to the process of humans annotating messages as ``\textit{human lie annotations}.'' 
% %
% This annotation method closely follows the approach described by \citet{peskov2020takes}.
%
While \citet{meta2022human} asked \textit{ex post facto} if any opponents were a computer, we inform players before play that there is a computer and we ask human players their guess of the humanity of each opposing power. 
%
% Our methodology for
% annotating human lies is closely aligned with the approach described
% by \citet{peskov2020takes}.
%
%\textbf{First}, humans must input a set of moves at the beginning of each turn, allowing us to capture human intent and hence persuasion (Section~\ref{sec:persuasion-detection}).
%
%allowing us to quantify deception (Section~ \ref{sec:deception-detection}). \textbf{Third},   for us to answer the last question.
%\footnote{We also require the players to adjust their stance using a 5-point Likert scale to indicate any friendliness or hostility and ask them to fill out a survey after each game (details are in Appendix~\ref{sec:survey_details}).} 

%RESOLVED \jbgcomment{below paragraph has formatting issues: spell out numbers less than 10 (I prefer 100), use en dash for ranges.}

%To investigate how \cicero{} interacts with humans, we invite players to play Diplomacy with \cicero{} and collect data throughout the games.
%
There are two to four human players per game, totaling 69 over all 24 games.\footnote{
We recruit players from Diplomacy forums and we pay at least \$70 per game, which lasts approximately three hours. We do not collect demographic information.
}
Games typically finish after fourteen movement turns, where each movement turns is limited to a brisk ten minutes. 
There are two to four human players per game, and \cicero{} fills any remaining slots. 
%
% We run \cicero{} using \abr{nvidia} A100 \abr{pcie}PCIE 40GB, where each \cicero{} requires at least one GPU.
%
The game setup differs from Meta's \cicero{} study: players in this study know \emph{a priori} that they are playing a bot.
In total, we collect 27,665 messages from communication between humans
and \cicero{} (Table~\ref{tab:overall_stat}).

\begin{table}
\centering
\resizebox{\columnwidth}{!}{%
\begin{tabular}{cccccccc}
 \hline
  & \textbf{AUS} & \textbf{ENG} & \textbf{FRA} & \textbf{GER} & \textbf{ITA} & \textbf{RUS} & \textbf{TUR} \\
 \hline
 \textbf{Human}&1.0&2.4&6.7&4.7&3.9&3.3&1.1\\
 \textbf{\cicero{}}&7.9&3.8&7.7&6.3&4.1&5.5&6.9\\
 \hline
\end{tabular}
}
\caption{\cicero{} strategically plays Diplomacy better than humans, where humans have fewer supply centers compared to \cicero{} when playing with the same power assignments. We calculate the number of supply centers by the end of the game by averaging the results for human players and \cicero{}.}
\label{tab:scs}
\end{table}

\textbf{\cicero{} nearly always wins.}
Of twenty-four games, \cicero{} won twenty (84\%), which strongly suggests that \cicero{} has super-human strategy.
%RESOLVED \jbgcomment{let's be careful with language.  I don't think it proves it, but it is strong evidence}
%
On average, \cicero{} has more supply centers than human players by the end of the game (Table~\ref{tab:scs}). 
Humans are about as good as \cicero{} when playing powers that require careful coordination of actions, such as Italy, which needs to manage both fleets and armies. 
However, when playing powers that require less coordination, such as Austria with its limited coastline, the gap in supply center counts between human players and \cicero{} is larger (see breakdown by power in Appendix Figure~\ref{fig:faceted_scs}); England is the only power where Cicero's average supply center count does not increase.
%
% As turn goes by, \cicero{}'s supply center
% count rises for every power except England (), suggesting a powerful strategic engine.

%
%RESOLVED \jbgcomment{This is confusing, unpack}
%
 %The end-of-game supply center counts are shown in table \ref{tab:scs}.

% \textbf{Previous human-human experience do not necessarily indicate better performance in human-cicero games.} We use player's tournament statistics, combined with self-accessed ratings using a 5-point Likert scale in the survey to determine the player ratings. We measure player performances by comparing their end-of-game supply centers and the human average. Players with previous tournament experience perform better than their inexperienced counterparts. However, top players, including national champions and tournament finalists, do not perform better than merely good players.

\jbgcomment{Use consistent typography for F-score}
\wwcomment{edited.}

\begin{figure}[t!]
\includegraphics[width=0.45\textwidth]{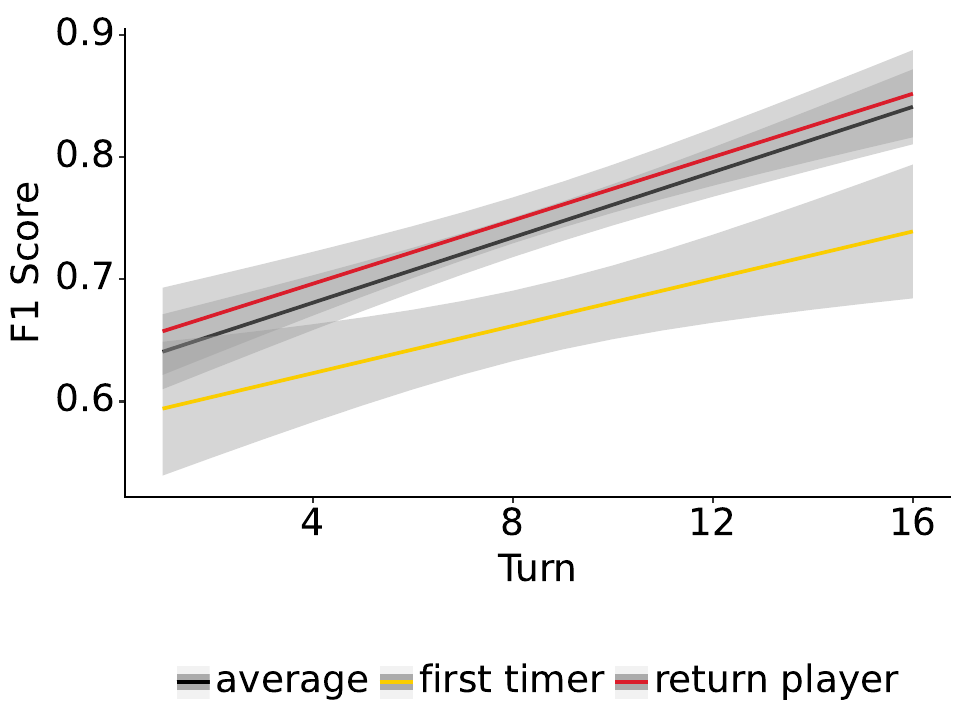}
\caption{Returning players (those who previously played against \cicero{} at least once) are better at correctly identifying other players as \cicero{} compared to first-time players. $F$-scores are presented for first-timers, returning players, and the average for all players, with smoothing via local regression \cite{doi:10.1080/01621459.1979.10481038}.}
\label{fig:image1}
\end{figure}

\textbf{Human players can reliably (but not perfectly) identify the bot. }
We calculate the average $F$-score of identification by turn (Figure~\ref{fig:image1}). 
By the end of the first movement turn, human players have an average $F$-score of 0.58, which keeps increasing until the end of the game. 
At game end, the average $F$-score is 0.81. 
Even for players in their first game against \cicero{}, the average $F$-score reaches 0.77. 
Players who previously played against \cicero{} at least once are better at identifying it. 
This suggests that \cicero{} can no longer pass as human 
% (as in~\cite{meta2022human}) 
once humans are aware of the possible existence of such agents.
% do we talk about heuristics of detecting cicero?

%RESOLVED \jbgcomment{The normalization is important and needs to be explained more clearly and justified.}
\subsection{Lies annotation}
\label{sec:result_human_lie}

This section analyzes players' \textit{deliberate lies} in sent messages and \textit{perceived lies} in received messages.
Because \cicero{} sends more messages than humans, we normalize \textit{perceived lies} by the number of messages that humans receive from \cicero{} and humans (6,960 and 2,276), while we normalize results of \textit{deliberate lies} by the number of total messages that humans send.

\jbgcomment{Need to define the normalization more rigorously}
\wwcomment{I provide more specific normalization definition}

\textbf{Humans feel that \cicero{} lies more often. }
%
% Of 9,236 messages received by human players, 1,167 (12.6\%) are perceived as lies.
%
Humans perceive 14.4\% of the 6,960 messages they receive from \cicero{} as lies (which is 1,005 messages, Figure~\ref{fig:perceived_lies}). In contrast, they perceive only 7.1\% of the messages from other humans as lies (which is 162 out of 2,276 messages).
%
% To obtain the unbiased frequency of lie perception, we normalize the data by the number of messages humans receive.
%
% Still, \cicero{} is perceived as
% sending lies twice as often as humans .
%
%
In the survey (detailed in Appendix~\ref{sec:survey_details}), players also think humans communicate more transparently than \cicero{}.
% Players believe \cicero{} is more deceptive than human players.
%
However, humans are not good at detecting lies.
Within 2,276 Human-Human messages; humans
can correctly identify five lies (0.2\%), suggesting a small overlap
between actual lies and perceived lies.
%
% In other words, players are not very good at detecting lies from other
% human players and they often wrongly classify truths as lies.

\jbgcomment{throughout the paper, there are several instances where
  you have ``X in figure Y''.  You can just talk about the specific
  results and have a parenthetical: Z\% of participants idintified
  \cicero{} (Figure A).}
\wwcomment{Feng and I have replaced with (Figure \dots)}
% We compare deception and persuasion between humans and \cicero{}. 
% %
% Though we introduced deception and persuasion functions for quantitative evaluation, we provide samples of humans and \cicero{} messages for a qualitative perspective.%the qualitative evaluation is not ignored and is shown with 
%
%
\begin{figure}[t]
    \centering
    \includegraphics[width=0.45\textwidth]{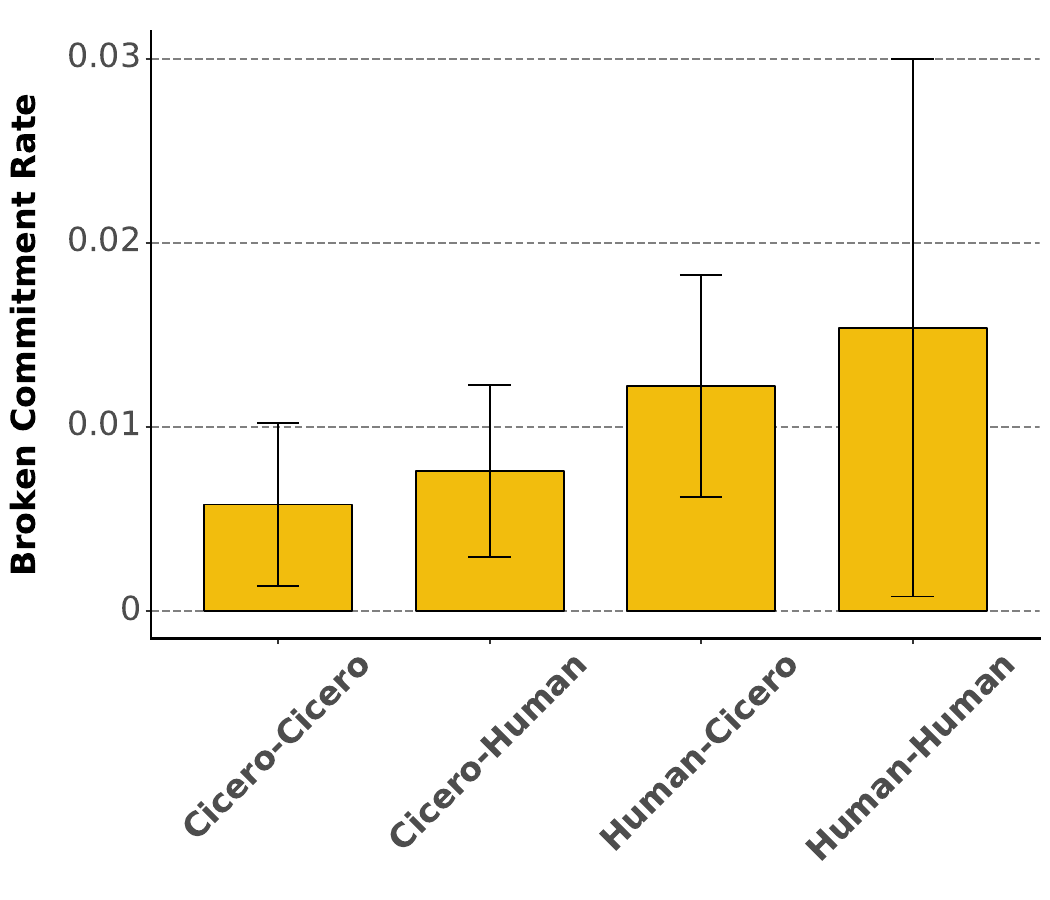}
    \caption{Though \cicero{} was perceived to lie more, we detect more broken commitments from humans.
    Each bar chart is the broken commitment rate (Equation~\ref{eq:bc}) labeled by sender and reciever: \cicero{} breaks commitment with \cicero{}, \cicero{} breaks commitment with Human, Human breaks commitment with \cicero{}, and Human breaks commitment with Human. Error bars represent $\pm$ one standard deviation over twenty-four games. }
    \jbgcomment{This phrasing is awkward: this is about detection
      vs. perception, make this clearer (although \cicero{} was perceived
      to lie more, we detected more human lies).}
    \wwcomment{edited! thank you}
    \label{fig:dec_rate}
\end{figure}
% \subsection{Quantitative Evaluation}
% We separate player groups into four combinations of pairs: \texttt{Human-Human}, \texttt{Human-\cicero{}}, \texttt{\cicero{}-Human}, and \texttt{\cicero{}-\cicero{}}.
% %
% We evaluate deception and persuasion by counting potential messages that players send out.
% %
% We keep track of the sender and the receiver.
% %
% %We count a message if a message's sender and recipient belong to both of a pair of player groups.
% %
% For example, France (Human) tells Germany (\cicero{}) to leave Burgundy. 
% %
% This will be counted towards the \texttt{Human-\cicero{}} group. 
% %
% The denominator is the number of messages that the first player of the group sends out, e.g., persuasion rate of \texttt{Human-\cicero{}} can be derived as $\frac{\text{total persuasive messages that a Human sent to \cicero{}}}{\text{all messages that a Human sent to \cicero{}}}$.

\textbf{Humans return the favor by saying they lie to \cicero{} more often.} 
%
% Not only do humans see \cicero{} as somewhat lying, but they also
% like to lie to \cicero{} (Figure~\ref{fig:human_lies}).
%
% To ensure unbiased in \cicero{} and humans, we normalize these deliberate lies by the number of messages humans send. 
%
Over 7,395 messages that
humans sent out, 273 of these are purposeful lies to \cicero{} (3.7\%),
while there are only forty-five lie messages to other human players (0.6\%).
This reflects that humans strategically lie more often to \cicero{} while believing that \cicero{} does not hold  grudges.

% \cicero{} resists lies from humans, or human lies have
% a low effect on \cicero{}'s decision-making.

\subsection{Detection}
\label{sec:result_detection}

After validating our automatic metrics, we compare human and computer deception and persuasion.

% We show detection accuracy following our definitions (Section~\ref{sec:detection}) and results for deception and persuasion. Note that we normalize the number of detected deception and persuasion by the number of total messages that occur in a certain group.
%

\textbf{Our broken commitment and persuasion detection is relatively effective.} 
To ensure that our detection is good enough, we sample around 4800 messages for an accuracy study (Table~\ref{tab:expert_detect}).
Broken commitment detection has a \textit{precision} of 0.51 and a \textit{recall} of 0.71.
% , demonstrating that it is quite effective.
%
Our precision is lower than our expectation due to errors in parsing a complex English to \abr{amr} \emph{and} a definition that only detects commitments at a move level (Appendix~\ref{sec:deception_limitation}). 
The broken commitment can \textbf{only} detect when a move in a message $A^{i\to j}_{\text{msg}}$ and a final move$A^{i\to j}_{\text{final}}$ are not aligned.
There are examples that  cannot detect, e.g. an agreement
to an alliance (Table~\ref{fig:alliance}) or a long conversation
before committing a deception (Table~\ref{fig:longconvo_dec}).
%
% We hand label samples and avoid using human lie annotation as a \textit{ground truth} of our deception since there is a message that a human player sees as \textit{Truth}, though they
% decide to lie later (Figure~\ref{fig:lie_annotation}) and this is detected by our deception detection.
%
Accuracy for persuasion is better;
precision rises to 0.81, and recall to 0.72.
%
% This shows that our detection is trustworthy and able to detect good
% number of broken commitment and persuasion acts.

\jbgcomment{Make it clear which of these are errors and which are
  problems of definition.  It might be good to include these
  judgements in the appendix.  (So you can refer to explicit
  examples)}
\wwcomment{I redirect to Appendix~\ref{sec:deception_limitation} providing three broken commitment events that we observe from our detection and human lies}

\textbf{Broken commitments are inconsistent with the perceived lie annotations}.
Humans break commitments more frequently than \cicero{} (Figure~\ref{fig:dec_rate}):
Humans break commitments with \cicero{} 1.2\% of the time
(63 out of Human--\cicero{} 5,151 messages) and do so to other human players 1.5\% of the time (35 out of Human--Human 2,276 messages).
On the other hand, \cicero{} breaks commitments at a lower, consistent rate, deceiving humans 0.76\% of the time and \cicero{}  0.57\% of the time (53 out of 6,960 messages and 77 out of 13,319 messages, respectively).
\begin{table}[t]
    \begin{tabular}{p{1.2cm}p{5.5cm}}
    \hline
    \textbf{Sender} & \textbf{Message}   \\ 
    \hline
     Turkey & \small{Hey Italy! I think the I/T is the strongest alliance in the game, would you be interested in working together} \\ 
    \rowcolor{grayish}  % Coloring a specific row
    Italy & \small{Of course! As long as you don't build too many fleets, I'm open to working with you against austria!} \\
    \end{tabular}
    \caption{The broken commitment detector ($\text{BC}(\cdot)$) has its limitation where it cannot capture deception in alliance agreement when Italy (human) deceives Turkey (\cicero{}).}
    \label{fig:alliance}
\end{table}
%
% \begin{figure}
%     \centering
%     \includegraphics[width=0.45\textwidth]{figures/alliance_agreement.pdf}
%     \caption{The broken commitment detector ($\text{BC}(\cdot)$) has its limitation where it cannot capture deception in alliance agreement when Italy (human) deceives Turkey (\cicero{}).}
%     \label{fig:alliance}
% \end{figure}
%
% This shows that deception in humans is hard to capture, though it is possible if the conversation contains pieces of information that we can cross-reference to the world state, which in this work, is a Diplomacy game state.

\begin{figure}
    \centering
    \includegraphics[width=0.45\textwidth]{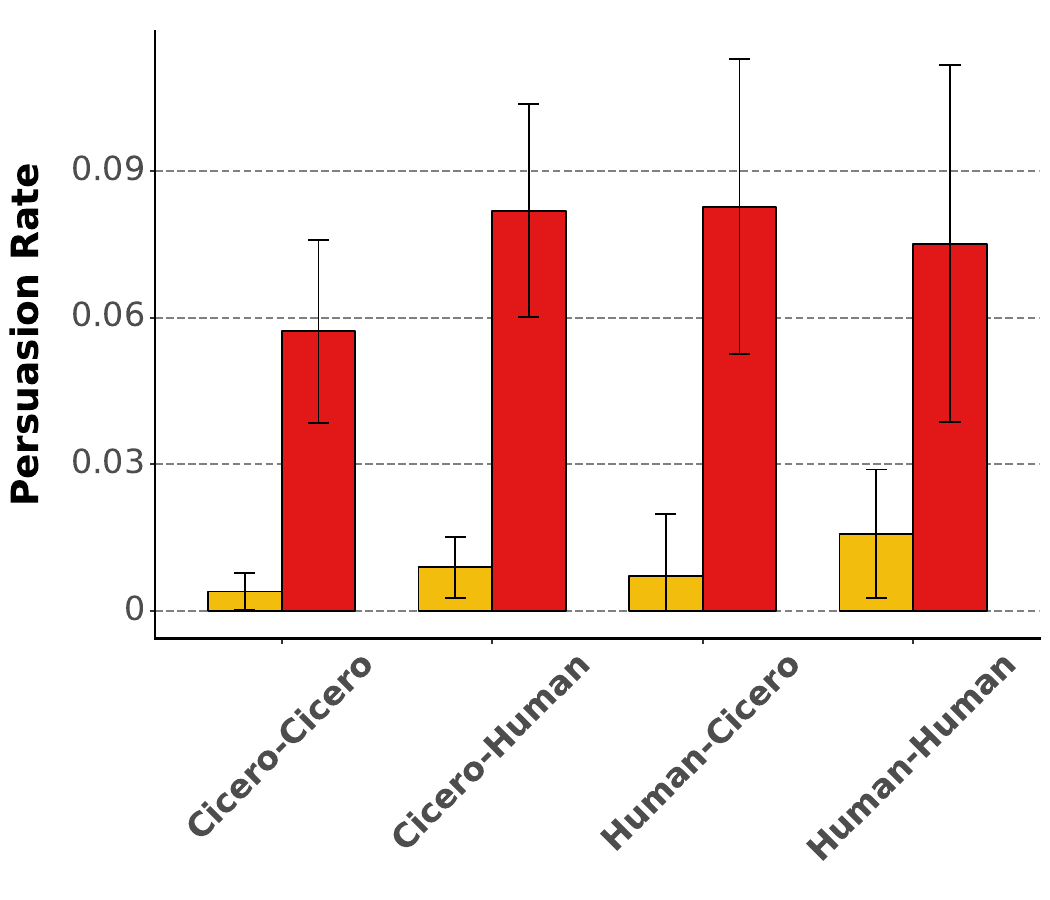}
    \caption{Humans outpace \cicero{} in persuasive effectiveness. Humans have a higher persuasion success rate, which we measure by comparing the number of successful persuasions (yellow, left) to the total number of persuasion attempts (red, right). We analyze success rates across four groups: \cicero{} persuades \cicero{}, \cicero{} persuades Human, Human persuades \cicero{}, and Human persuades Human. Error bars represent the $\pm$~one standard deviation range from the aggregate of interactions in 24 games.}
    \label{fig:all_persuasion}
\end{figure}

\textbf{Humans are more persuasive.}  
For persuasion to happen, we need first an \emph{attempt}, initiated by a sender, and then \emph{success} when the receiver adopts the suggestion.
%
% Humans both make more attempts and succeed more often: 
% clearly have more potential in persuasion and have higher success than \cicero{} overall (yellow bars in Figure~\ref{fig:all_persuasion}). 
%
% Of 7,427 messages, humans successfully persuaded another human player by 36 messages and successfully persuaded \cicero{} by 37 messages, which is 0.48\% and 0.50\%.
% %
% Though \cicero{} is not quite as persuasive as humans, among 20,270 messages, \cicero{} can persuade humans with 62 messages (0.31\%) and persuade other \cicero{}s with 53 messages (0.26\%). 
%
Both humans and \cicero{} on a per-message basis\footnote{Although because \cicero{} communicates more overall, humans attempt more times per game.} try to persuade at the same rate (around 8\% of the time, per Figure~\ref{fig:all_persuasion}).   
The success rate of human persuasion is 21.1\%
%(36 out of 171 attempts)
at persuading other humans and 8.6\% 
% (37 out of 426 attempts) 
at persuading \cicero{}. 
\cicero{} is less persuasive; its success rate is only 10.9\% in persuading humans and 7.0\% in persuading other bots.
%
% This shows that humans are more cooperative in games when persuading other players and when asked or suggested to take action. 

In summary, humans are more deceptive and more persuasive than \cicero{}. 
Detection is possible, but defining a sequence of conversations as persuasion or deception is still difficult. 
Our reported numbers are low because both humans and \cicero{} engage in extensive back-and-forth discussions before making moves that can be definitively classified as persuasion or deception.

% \begin{figure}[t]
%     \centering
%     \includegraphics[width=0.45\textwidth]{figures/success_persuasion.pdf}
%     \caption{Success Persuasion using $Per()$ function by groups of players.}
%     \label{fig:success_per}
% \end{figure}

\jbgcomment{Should we distinguish the human lie annotations from our detections?  Is this making the story too complicated?}
\wwcomment{If the subsection with the introduction makes it more clear, then this current version could work. please let me know what you think.}

% \subsection{Qualitative evaluation} 

\section{Related Work}
\label{sec:related}

Large language models are becoming ubiquitous in many tasks: fact-checking~\citep{lee2020language,lee2021towards}, text generation~\citep{devlin2018bert,brown2020language, touvron2023llama} including coding~\citep{roziere2023code}.
All of these tasks require users to trust models' outputs.
However, models are \emph{not} always reliable; they could produce hallucinations or conflict with established facts~\citep{Ji2023nlg,zhang2023siren,si2023large,yao2023llm}.
To mitigate this, their outputs often need to be verified against datasets~\citep{Thorne2018FEVERAL,wadden-etal-2020-fact,Schuster2021GetYV,Guo2021ASO}.
Studies have used adversarial examples to expose weaknesses and to raise awareness~\citep{eisenschlos2021fool,schulhoff2023ignore,liu2023prompt,lucas2023fighting} 
To address the issue of unreliability, controllable LMs been proposed by having steps to inject facts for better reasoning~\citep{adolphs2021reason}, or by prompting techniques, such as chain-of-thought prompting, to enhance reasoning abilities~\citep{wei2022chain,wu2022aichains}. 
Moreover, some studies focus on \abr{ai}-Generated misinformation~\citep{zhou2023synlies}, probing model to understand internal states when LLM utters truthful or false information~\citep{azaria2023internal,li2024inference}.

% \textbf{Deception and Persuasion in Game-based communication.} 
%intro to general communication (not game yet)
% Research has focused on human-AI cooperation, evaluating conversational agents in cooperative games \citep{chattopadhyay2017evaluating}, understanding through game theory and reinforcement learning \citep{schelble2021understanding}, and examining social perceptions of AI partners \citep{ashktorab2020human}. 
%\textbf{AI with social abilities.} 
%
Deception and persuasion are studied within social contexts.
% , encompassing a broad range of research areas. \citet{wang2019persuasion} collect persuasion strategies and analyze their effectiveness in generating donations. 
%
\citet{huang2023artificial}'s meta-analysis concludes \abr{ai} can match humans in persuasion,
and \citet{deck2023bullshit} attributes some of the success to the ability to generate ``bullshit'', 
which are part of applications in marketing and public relations~\citet{hallahan2007defining}.

Part of what makes games like \textit{Diplomacy} as an object of study appealing is the ongoing race between humans and computers in games~\citep{kim2018performance}; 
initial work on the language of \textit{Diplomacy}~\citep{Niculae:Kumar:Boyd-Graber:Danescu-Niculescu-Mizil-2015} unlocked follow-on work both in \textit{Diplomacy}'s agreements~\citep{kramar2022negotiation} and in other games such as ``The Resistance: Avalon''~\citep{light2023avalonbench,xu2023language, stepputtis2023long,lan2023llm} and ``Mafia''~\citep{ibraheem2022putting}.
%
% Our research aligns closely with studies on lies ~\citep{peskov2020takes}, betrayal , and  within the context of the game Diplomacy. 
%
% Although~\citet{meta2022human} did not explicitly attribute deception capabilities to \cicero{}, our observations of deceitful behavior in \cicero{} have inspired us to benchmark its abilities in deception and persuasion.

%\textbf{Humans and Language Models.} 

\section{Conclusion and Future Work}
\label{sec:conc}

Our research confirms that \cicero{} can win most games of \textit{Diplomacy}, but has not \emph{mastered} the nuances of communication and persuasion.
Truly mastering the game requires systems that (a) can maintain
consistency between their communication and actions, (b) can
communicate at a variety of levels, including tactics, strategy, and
alliances, and (c) can use communication as a tool of persuasion,
deception, and negotiation.

\textit{Diplomacy} remains an attractive testbed for communication and strategic research.
It offers the ability to build more comprehensive systems that understand relationship dynamics, can engage in realistic but hypothetical conversations, and that can be robust to the deceptions of others.
Because these are places where humans still outpace \abr{ai}, it also offers synergies for developing human--computer collaboration.

And while these tasks are important withing the silly game of \textit{Diplomacy}, they can help solve long-standing \abr{ai} problems:
helping users deal with \abr{llm}-generated deception,
collaborating with users on grounded planning,
and understanding human norms of reciprocity, cooperation, and communication.
This will help \abr{ai} not just be fun for negotiation in board games but safer and more trustworthy when we negotiate everyday problems.

\clearpage
\section*{Limitations}
\label{sec:limitation}

To gain a clearer understanding of cooperation and deception between human and \cicero{}, we need to experiment with different game setting and turn duration. For example, inexperienced players might be overwhelmed by the amount of communications in early movement turns; prolonging the turns to 15 minutes might improve communication quality. Furthermore, this study collects only 24 blitz games of human playing against \cicero{}. The power distribution of participants is imbalanced: the most frequent power---France---has 14 appearances, whereas the most underrepresented power, England, has only five. Class imbalance \cite{Fernández2018} could potentially impact the feature weights in our regression model for player performance.
% Joy: add limitations of detecting deception/cooperation

Since the \abr{AMR} parser does not always predict correct intentions, this has an effect on our precision and recall of deception and persuasion detection protocol.
Our detection cannot cover such long conversations that humans have; we limit detection to only checking back to the previous message, and this makes our detection miss cooperation, deception, and persuasion when humans and \cicero{} discuss the plan.

\section*{Ethical Considerations}
We recruited players from Diplomacy forums, including Diplomacy Discord and reddit.
We paid them over \$70 per three-hour game and did not collect demographic information.
Procedures in our study involving human subjects received \abr{irb} approval and are compliant with \abr{acl} Code of Ethics. Human participants are aware of the purpose of the study and are free to withdraw at any time. There are no potential risks or discomforts from participating. We obtained consent from all participants.

Researching how artificial intelligence (\abr{ai}) can deceive and persuade helps us understand its capabilities. This investigation reveals that AI can execute complex tasks effectively. However, it is important to note that these abilities do not significantly risk society.

\section*{Acknowledgements}

We thank Meta for granting access to over 40,000 games played on the online platform \url{webdiplomacy.net} and for open sourcing \cicero{}.
This commitment to open science allowed this independent reproduction of \cicero{}'s juggernaut abilities but also let us have some fun.
We especially thank to Mike Lewis for offering valuable insights into \cicero{}'s communication.

Our thanks also go to Tess Wood for training \abr{amr} annotators, Sarah Mosher for their English-\abr{amr} annotations, and Isabella Feng for her exploration of \abr{llm}-based \abr{amr} parsing. 
We thank Kartik Shenoy, Alex Hedges, Sander Schulhoff, Richard Zhu, Konstantine Kahadze, and Niruth Savin Bogahawatta for setting up \abr{daide} baselines.

We also thank the small community of researchers looking at communication and deception in Diplomacy for their feedback, commentary, and inspiration: Michael Czajkowski for discussing the nuances of detecting persuasion; Stephen Downes-Martin for teaching us that deception is \emph{far} more than lies; Karthik Narasimhan and Runzhe Yang for their insights into lie detection and stance; 
and Larry Birnbaum and Matt Speck for discussions on mapping \abr{daide} and English.
And thanks to Justin Drake, Niall Gaffney, and the members of \abr{tacc} for setting up environments for computer--computer games and making sure that we had \abr{gpu}s ready when players were ready to play.

Finally, sincere thanks to the member of the Diplomacy community who took the time to play against \cicero{} in this unconventional setting.

%
% We appreciate the contributions of Eddie Schoute, Matthew Totonchy, Ian Rudnick, Christopher Rawles, Ben Schroeder, Robert C. Schuppe, Andrew Guo, David Graff, Jack Henrichs, Kirk Vaughn, Connor.e. knight, Marko Papić, Christopher Ward, Sloth, Mikalis Kamaritis, Jordan Connors, Andrew Zicks, Tommy Anderson, Aashutosh Maheshwary, Dr. Abhishek Singhal, David Smith, and Parul Bansal for their participation in the Human-\cicero{} experiments and valuable feedback.
% Finally, we thank Meta for granting access to over 40,000 games played on the online platform \url{webdiplomacy.net} and also extend our special thanks to Mike Lewis for offering valuable insights into \cicero{}'s communication.
%
%
This material is based upon work supported by the Defense Advanced Research Projects Agency (DARPA) under Agreement Nos. HR00112290056 and HR00112490374. 
Any opinions, findings, conclusions, or recommendations expressed here are those of the authors and
do not necessarily reflect the view of the sponsors.

% CHIRON
% [ALLAN, CHIRON, …?]
% Tess, annotators
% Stephen 
% USC interns/ms students (jon needs to check)
% Sydney non-author students
% Princeton people (-Brandon)
% Game players
% TACC
% \cicero{} RFP data sharing

\bibliography{bib/journal-full,bib/anthology,bib/custom,bib/jbg}

\appendix
\renewcommand{\thefigure}{A\arabic{figure}}
\setcounter{figure}{0}
\renewcommand{\thetable}{A\arabic{table}}
\setcounter{table}{0}

\clearpage

\section{Background: Gunboat Diplomacy}
\label{sec:gunboat_background}

\citet{paquette-19} develop a Diplomacy interface and was the first to publish an agent trained by human data and trained through self-play using reinforcement learning with Advantage Actor-Critic (A2C)~\citep{mnih2016asynchronous}.
DeepMind employed policy iteration in their reinforcement learning training~\citep{anthony2020learning}, whereas Meta utilized a combination of regret matching, equilibrium search, and deep Nash value iteration~\citep{gray2020human,bakhtin2021no}.
The most recent advancement is by Meta~\citep{bakhtin2023mastering}, regularizing the agent's policy with human policy data. 
This strategic enhancement culminates in the development of \cicero{}~\citep{meta2022human}.

\section{AMR}
\label{sec:amr_appendix}

\subsection{AMR Annotation}
\label{sec:amr_annotation_appendix}
Building on general \abr{amr} annotation guidelines, we established additional Diplomacy-specific \abr{amr} annotation guidelines, including what and how to annotate. Unlike general \abr{amr} annotations, where all sentences are fully annotated, in Diplomacy \abr{amr} annotations, some utterances are only partially (or even not at all) annotated, based on the degree of usefulness for Diplomacy. 3,306 of 8,878 human-annotated utterances
contain partial information with underspecified units, locations, and countries. As Diplomacy
players often communicate sentences that lack
full details about the game state which they can infer from a visualized map. This directly shows in \abr{amr} with the missing object. We provide examples of underspecified utterances, missing unit location (Figure~\ref{fig:amr_loc}) and missing unit country (Figure~\ref{fig:amr_country}).
\begin{figure}[h]
    \small{ \texttt{(m / move-01\\
    \hspace*{4 mm}:ARG1 (u / unit\\
    \hspace*{8 mm}:mod (c2 / country\\ 
    \hspace*{16 mm}:name (n2 / name :op1 ``Austria'')))\\
    \hspace*{4 mm}:ARG2 (p2 / province\\ 
    \hspace*{8 mm}:name (n3 / name :op1 ``Brest'')))\\}}
    \caption{\label{fig:amr_loc}Parsing from English to \abr{AMR} can have underspecified utterances. The English text is from Austria talking to Italy, ``\textit{Let's work on our plan, I'm moving to Brest}''. We show an \abr{AMR} with missing unit location referencing from English text.}
\end{figure}

\begin{figure}[h]
    \small{ \texttt{(m / move-01\\
    \hspace*{4 mm}:ARG1 (u / unit\\
    \hspace*{8 mm}:location (p2 / province\\  
    \hspace*{16 mm}:name (n / name :op1 ``Romania'')))\\
    \hspace*{4 mm}:ARG2 (p3 / province\\ 
    \hspace*{8 mm}:name (n3 / name :op1 ``Bulgaria'')))\\}}
    \caption{\label{fig:amr_country}AMR being underspecified in unit country where it parses from English text, ``\textit{just bumping Bulgaria from Romania}''}
\end{figure}
% % https://www.isi.edu/~ulf/amr/lib/amr-dict-diplomacy.html

Our Diplomacy Appendix of the \abr{amr} Annotation Dictionary lists \abr{amr} concepts (e.g. \amr{betray-01}), their related English terms (e.g. \ betray, stab, traitor, treason), annotation examples, any corresponding DAIDE code, and notes. \abr{amr} concepts with DAIDE equivalents include \amr{ally-01}, \amr{build-01}, \amr{move-01}, and \amr{transport-01}. We analyzed player messages for additional concepts of high Diplomacy communication value, and extended the Diplomacy \abr{amr} vocabulary (compared to DAIDE) by including concepts such as \amr{attack-01}, \amr{betray-01}, \amr{defend-01}, \amr{expect-01}, \amr{fear-01}, \amr{have-03}, \amr{lie-08}, \amr{possible-01}, \amr{prevent-01}, \amr{tell-01}, \amr{threaten-01}, and \amr{warn-01}, as well as roles such as \amr{:purpose} and \amr{:condition}. This allows annotators to easily mark sentences, e.g. ``Russia is planning to take you out as soon as possible.'' would use the concept \amr{attack-01}. We also extended \abr{amr} guidelines to cover gaining/holding/losing provinces, especially support centers.
% Full vocabulary dictionary is shown in Table{}.

The general \abr{amr} Editor includes a Checker that performs a battery of tests to ensure well-formed and consistent AMRs. We extended the Checker for Diplomacy AMRs, e.g.\ to ensure that for a \amr{build-01}, the location is an argument of \amr{build-01} itself, rather than an argument of the army or fleet being built.

\abr{amr} covers more Diplomacy content than DAIDE, not only due to additional concepts such as \amr{betray-01}, but also because arguments are syntactically optional. Unlike DAIDE with its rigid positional argument structure, \abr{amr} can thus represent underspecified information such as units with missing type, location or nationality; or agreements with a missing object. \abr{amr} can also accommodate additional arguments compared to DAIDE, for example the source and target of a proposal.

Because not all information needed for annotation is available in the raw text, we offer annotators access to dialog partners (speaker, recipient), season (e.g.\ Spring 1901) and a map with current deployments (as available). 

\begin{figure}[t]
    \centering
    \includegraphics[width=0.5\textwidth]{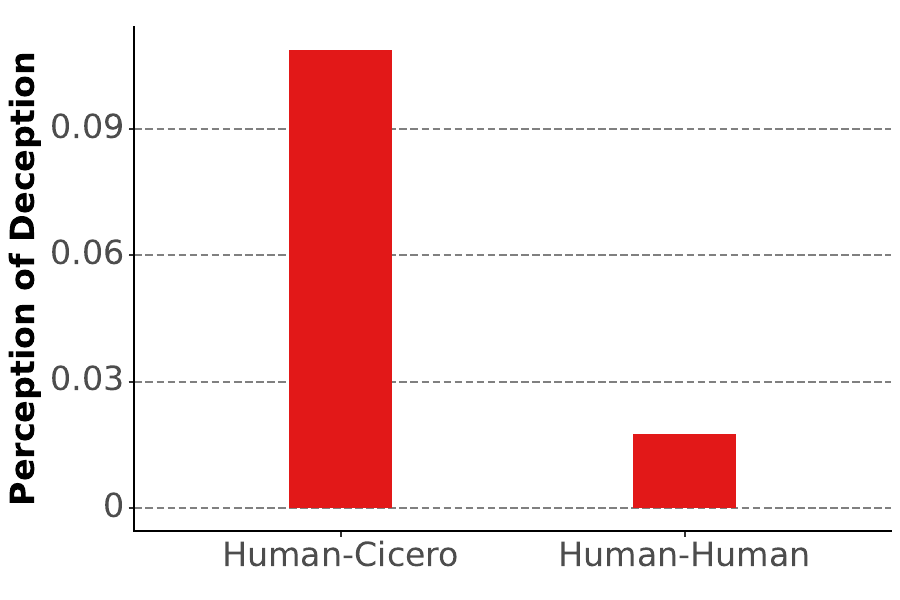}
    \caption{Perception of deception rate by human annotation.}
    \label{fig:perceived_lies}
\end{figure}
\begin{figure}[t]
    \centering
    \includegraphics[width=0.5\textwidth]{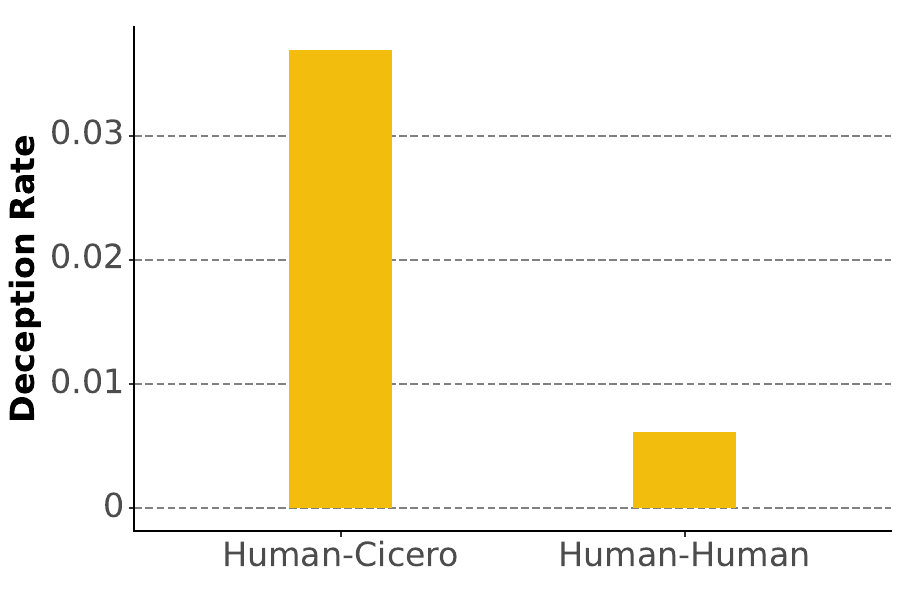}
    \caption{Deception rate by human self-annotation.}
    \label{fig:human_lies}
\end{figure}

\subsection{AMR Parser}
\label{sec:amr_parser_appendix}
While stylistic aspects play an indispensable role in sustaining engagement and interest among participants, factual information is more vital for informed decision-making. Detecting deception and persuasion in communications requires checking the relationship between message information and initial/final moves of a particular power. To address the nuanced challenge of distinguishing meaningful and informative content from stylistic dialogues in \textit{Diplomacy}, our focus herein is on developing a sophisticated pipeline for information extraction using \abr{amr} from text. We utilize a state-of-the-art Sequence-to-Sequence model from the Huggingface transformers library, fine-tuned with the AMR 3.0 dataset, for baseline semantic extraction. This approach facilitates the processing of \abr{amr} through amrlib, a Python module tailored for such tasks. The efficacy of our AMR parsers is assessed using the SMATCH score, the gold standard for evaluating \abr{amr} accuracy. We divided the annotated Diplomacy-AMR corpus into 5054 training, 1415 validation and 2354 test sentences and used similar parameters except for increasing the number of epochs from 16 to 32.

When fine-tuning our model for Diplomacy game communications, we shifted from the overly broad AMR 3.0 vocabulary to the tailored Diplomacy-AMR corpus introduced above, reducing irrelevant content and focusing on game-specific nuances. This strategic adjustment, alongside removing the original dataset to minimize bias, significantly improved our model's relevance and increased the SMATCH score from 22.8 to 61.9. 

We further enhanced accuracy through Data Augmentation, adding context to dialogues to aid the model's understanding of pronouns and strategic details, leading to a SMATCH score improvement from 61.9 to 64.6. Incorporating specific tokens for sender and recipient identities refined this approach, yielding additional gains from 64.6 to 65.4 in parsing accuracy.

By replacing (1) pronouns with country names and (2) some provinces in abbreviations with full names, we increases the SMATCH score to 66.6.

\subsection{Assessing the Role of Communication in \cicero{} vs. \cicero{} Games}
\label{sec:amr_CCgames_appendix}
We conduct 180 computer-computer games with 60 games for each communication level (Natural Language, Random Messages and \abr{amr} Information) and collected data to build a corpus for \cicero{}-\cicero{} Games. This corpus comprises instances of games where we record the power and communication assignments and the final scores (e.g. 'Game1': 'AUS 0, ENG 0, FRA 4, GER 10, ITA 5, RUS 6, TUR 9. (FRA GER TUR)' with the three powers shown in parentheses being identified as communicative). The communication strategies are randomly assigned to powers.  We regress the number of end-of-game supply centers on a dummy variable for the powers played (using Russia---the average player---as the baseline) and the communication strategy (using random messages as the baseline).  We plot the coefficients with classical 95\% confidence intervals. The effects of power selection are substantially larger than different communication strategies, none of which are significantly different from random messages at the $p<.05$ level.

\subsection{Future experiments: AMR information \cicero{}}
Since we have evidence that \cicero{}'s win weighs on its strategic rather than communication abilities (Section~\ref{sec:Comp_comp_games}).
To further study this, we want to downgrade \cicero{}'s communication and collect more human-\cicero{} games to see whether \cicero{} wins at the same rate (previously 84\% against humans). 
We conduct 5--10 games using the same setup as in the Human-\cicero{} games (Section~\ref{sec:human_comp_games}).
The only difference is \cicero{}.
We will limit \cicero{} communications from natural language to AMR information where it mostly captures \textit{move intent}.
\begin{table}
\centering
\begin{tabular}{ccrr}
\hline
& &\multicolumn{2}{c}{\textbf{Detection}} \\
\hline
& & TRUE & FALSE \\
\hline
\multirow{2}{*}{\textbf{Expert}} & TRUE &20 &8 \\
&FALSE &19 & 4745 \\
\hline
\end{tabular}
\caption{Total 4,792 messages (from Human/\cicero{} to Human/\cicero{}) comparing TRUE/FALSE whether expert humans see as a lie and whether detected as a broken commitment by our detection.}
\label{tab:expert_detect}
\end{table}

% \begin{table}[t]
% \centering
% \begin{tabular}{cccc}
% \hline
% & &\multicolumn{2}{c}{\textbf{Detection}} \\
% \hline
% & & TRUE & FALSE \\
% \hline
% \multirow{ 2}{*}{\textbf{Lie Annotation}} & TRUE &3 &72 \\
% &FALSE &22 & 1514 \\
% \hline
% \end{tabular}
% \caption{Total 1611 human send-out messages comparing TRUE/FALSE in human lie annotation and in broken commitment detection.}
% \label{tab:ann_detect}
% \end{table}

\begin{table}
\centering
\begin{tabular}{ccrr}
\hline
& &\multicolumn{2}{c}{\textbf{Expert}} \\
\hline
& & TRUE & FALSE \\
\hline
\multirow{ 2}{*}{\textbf{Lie Annotation}} & TRUE &3 &72 \\
&FALSE &13 & 1523 \\
\hline
\end{tabular}
\caption{Total 1,611 human send-out messages comparing TRUE/FALSE in human lie annotation and in expert hand labeling.}
\label{tab:ann_expert}
\end{table}

% \begin{table}[t]
% \centering
% \begin{tabular}{cccc}
% \hline
% & &\multicolumn{2}{c}{\textbf{Detection}} \\
% \hline
% & & TRUE & FALSE \\
% \hline
% \textbf{Perceived} & TRUE &6 &283 \\
% \textbf{Lie Annotation}&FALSE &16 & 1563 \\
% \hline
% \end{tabular}
% \caption{Total 1868 humans received messages comparing TRUE/FALSE whether humans perceived as a lie and whether detected by our broken commitment.}
% \label{tab:plie_detect}
% \end{table}

\begin{table}[t]
\centering
\begin{tabular}{ccrr}
\hline
& &\multicolumn{2}{c}{\textbf{Expert}} \\
\hline
& & TRUE & FALSE \\
\hline
\textbf{Perceived} & TRUE &5 &284 \\
\textbf{Lie Annotation}&FALSE &7 & 1572 \\
\hline
\end{tabular}
\caption{Total 1,868 humans received messages comparing TRUE/FALSE whether humans perceived as a lie and whether human experts see as a lie.}
\label{tab:plie_expert}
\end{table}

\section{Deception detection limitations}
\label{sec:deception_limitation}
\wwcomment{As Jordan suggested, we should clearly point out errors (wrong detect) and problems of definition (cannot detect).}
We want to discuss deception detection further here to state errors and limitations. Since we mentioned our \textit{precision} for deception detection is quite low (Section~\ref{sec:result_detection}), we hereby expand on detection limitations and also compare to human (deliberate) lies as follows:
\begin{enumerate}
    \item what our detection is likely to miss when humans lie,
    \item what our detection mistakenly detects as deception,
    \item what humans annotate as \textit{Truth}, though it is a \textit{break of commitment} and our detection can detect correctly.
\end{enumerate}

\textbf{Humans often lie about relationships.} Detecting broken commitment at the relationship level is not possible for our detection (Table~\ref{fig:alliance} and Table~\ref{fig:alliance_2}).
This is a limitation of our deception definition, which focuses on moves.
Though it is possible to extract the relationship among players to see conflicts in the messages, we avoid doing so because the relationship is another topic to study in more detail.
At this stage of our work, we cannot train a model predicting relationships that can be circulated from game states, dialogue, and moves without collecting human data first.
Therefore, we have relationship tracking from human players for a study in the future.

\textbf{AMR limits broken commitment detection \textit{precision}.} Some messages are parsed incorrectly, which can be seen as a \textit{commitment is broken} (Table~\ref{fig:invalid_amr}). 
This makes the detection falsely detect \textit{truthful} messages as deceptive (increases false positive examples which decreases \textit{precision}).
Another limitation we observed is when one accepts the proposal but does not follow as commit using a short answer, e.g. \textit{Yes, I agree.} or \textit{Sure}.
Our \abr{amr} parser sometimes hallucinates and extracts invalid moves, which can be mistakenly detected as breaking a commitment. 
\begin{table}
    \begin{tabular}{p{1.2cm}p{5.5cm}}
    \hline
    \textbf{Sender} & \textbf{Message}   \\ 
    \hline
     Austria & \small{That's an interesting opening. Was the bounce in EC planned?} \\ 
      Austria & \small{Do you think Germany will work with you against France?} \\ 
    \rowcolor{grayish}  % Coloring a specific row
    England & \small{Yeah it would be great if we team up} \\
    \end{tabular}
    \caption{The broken commitment detector ($\text{BC}(\cdot)$) cannot detect deception in alliance agreement when Austria (human) deceives England.}
    \label{fig:alliance_2}
\end{table}

\begin{table}[t]
    \begin{tabular}{p{1.2cm}p{5.5cm}}
    \hline
    \textbf{Sender} & \textbf{Message}   \\ 
    \hline
     Turkey & \small{If you retreat from Serbia into Budapest, then I'm in} \\ 
    \rowcolor{grayish}  % Coloring a specific row
    Italy & \small{I will do that if Serbia gets dislodged} \\
    \end{tabular}
    \caption{Italy agrees with the condition that the Turkey unit should move out of Serbia; however, our \abr{amr} parser captures Italy's sentence as \textit{``I will move to Serbia,''} which is invalid and makes our detection detects deceptive when Italy does not move to Serbia.}
    \label{fig:invalid_amr}
\end{table}

\begin{table}[t]
    \begin{tabular}{p{1.2cm}p{5.5cm}}
    \hline
    \textbf{Sender} & \textbf{Message}   \\ 
    \hline
     Germany & \small{Also, can we keep Burgundy clear?} \\ 
    \rowcolor{grayish}  % Coloring a specific row
    France & \small{Yes, we can do that. Are you moving to Helgoland?} \\
    \end{tabular}
    \caption{France (human) annotated \textit{``Yes, we can do that.''} as \textbf{Truth}, which contradicts the final move where France moves to Burgundy. This is captured as a broken commitment by the $\text{BC}(\cdot)$ function.}
    \label{fig:lie_annotation}
\end{table}

\begin{table}[t]
    \begin{tabular}{p{1.2cm}p{5.5cm}}
    \hline
    \textbf{Sender} & \textbf{Message}   \\ 
    \hline
     Germany & \small{I am going to try to move to English Channel} \\ 
    \rowcolor{grayish}  % Coloring a specific row
    England & \small{Sure} \\
     Germany & \small{It might help you hold London} \\ 
    \rowcolor{grayish}  % Coloring a specific row
    England & \small{Yeah I am holding London} \\
    \end{tabular}
    \caption{England (human) annotate \textit{``Yeah I am holding London''} as \textbf{Truth}, which contradicts the final move where an army in London moves to Edinburgh. This is captured as a broken commitment by the $\text{BC}(\cdot)$ function.}
    \label{fig:lie_annotation_2}
\end{table}
% \begin{figure}[t]
%     \centering
%     \includegraphics[width=0.45\textwidth]{figures/alliance_agreement_2.pdf}
%     \caption{The deception detector ($\text{Dec}(\cdot)$) cannot capture deception in alliance agreement when Austria (human) deceives.}
%     \label{fig:alliance_2}
% \end{figure}

% \begin{figure}[t]
%     \centering
%     \includegraphics[width=0.45\textwidth]{figures/deception_invalid_amr.pdf}
%     \caption{Italy agrees with the condition that the Turkey unit should move out of Serbia; however, our \abr{amr} parser captures Italy's sentence as \textit{``I will move to Serbia,''} which is invalid and makes our detection detects deceptive when Italy does not move to Serbia.}
%     \label{fig:invalid_amr}
% \end{figure}

% \begin{figure}[t]
%     \centering
%     \includegraphics[width=0.45\textwidth]{figures/human_annotation_truth.pdf}
%     \caption{France (human) annotated \textit{``Yes, we can do that.''} as \textbf{Truth}, which contradicts the final move where France moves to Burgundy. This is captured as deception by the $\text{Dec}(\cdot)$ function.}
%     \label{fig:lie_annotation}
% \end{figure}

% \begin{figure}[t]
%     \centering
%     \includegraphics[width=0.45\textwidth]{figures/human_annotation_truth_2.pdf}
%     \caption{England (human) annotate \textit{``Yeah I am holding London''} as \textbf{Truth}, which contradicts the final move where an army in London moves to Edinburgh. This is captured as deception by the $\text{Dec}(\cdot)$ function.}
%     \label{fig:lie_annotation_2}
% \end{figure}

\textbf{Human lie annotation is not always correct.} It is true that we have human annotations, and they can be seen as \textit{ground truth}. However, we sample annotations from four games data and comparing to expert labeling (lies in Table \ref{tab:ann_expert} and perceived as lies in \ref{tab:plie_expert}). This shows that humans are not good at predicting lies, and sometimes they are \textit{honest} but then decide to break their words later.
There are examples where humans commit to such action but do not follow, though they firstly annotate as a \textit{truthful} message (Table~\ref{fig:lie_annotation} and Table~\ref{fig:lie_annotation_2}).

\section{Survey Details}
\label{sec:survey_details}
The survey consists of 5-point Likert scale questions and free-form text questions. The questions are designed to measure the human players' perception of \cicero{}'s communication and their experience playing with \cicero{}. We also included questions to measure the players' expereince with Diplomacy and their general impression of \cicero{} for qualitative analysis. Overall, players believe that human communicate more transparently and are more strategically cooperative. Survey results are shown in table~\ref{tab:distribution}.

\begin{table*}
\centering
\begin{tabular}{l | ccccc | c}
 \toprule
 \textbf{Statement} & \multicolumn{5}{c|}{\textbf{Likert Scale (\%)}} & \multicolumn{1}{c}{\textbf{Num.}} \\
  & \textbf{1} & \textbf{2} & \textbf{3} & \textbf{4} & \textbf{5} & \textbf{Responses}  \\
  \midrule
 I am really good at Diplomacy.&0&8.3&25&41.7&9&25\\
 I am able to identify all AIs.&9.5&23.8&38.1&16.7&11.9&42\\
 I enjoy talking with the AIs.&14.3&38.1&33.3&7.1&7.1&42\\
 I was able to make plans with other players.&7.1&23.8&35.7&14.3&19&42 \\
 I was able to make plans with the AIs.& 21.4 & 31 & 19 & 19 & 9.5 &42\\
 Human players communicated transparently.& 7.1 & 14.3 & 33.3 & 35.7 & 9.5 &42 \\
 AI players communicated transparently.& 11.9 & 26.2 & 45.2 & 9.5 & 7.1 &42 \\
 \hline
\end{tabular}
\caption{Statements in the survey and their respective responses. Larger number in the Likert scale indicates more agreement.}
\label{tab:distribution}
\end{table*}

% \section{Supply Center Count of \cicero{} and Human by Turn}

\begin{figure*}
    \centering
    \includegraphics[width=\textwidth]{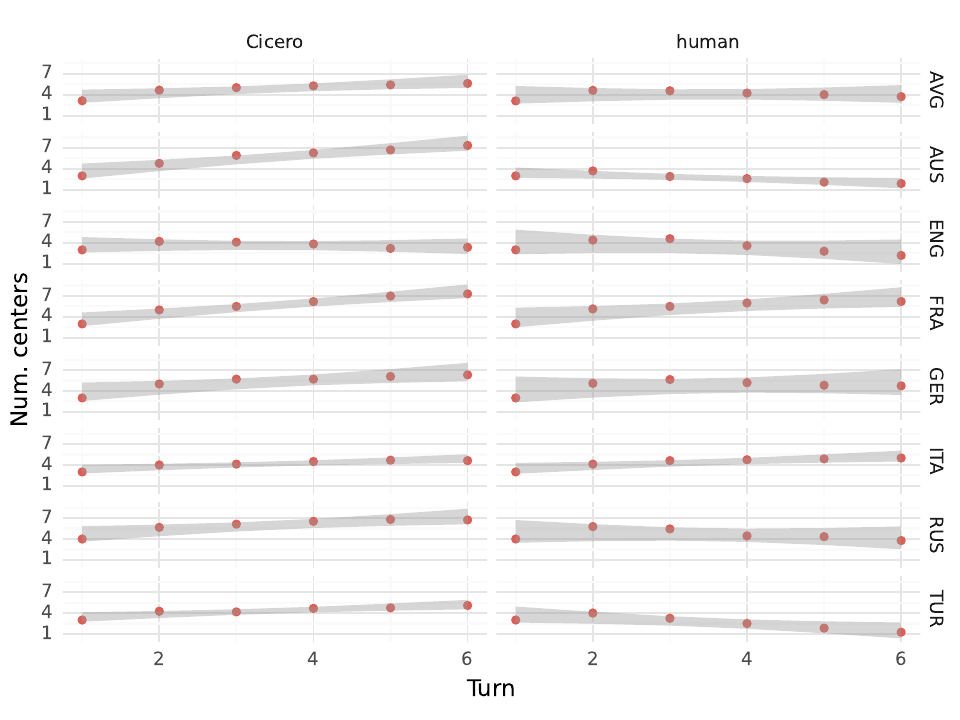}
    \caption{The average human player loses to \cicero{}. Human loses supply centers as game progresses unless playing as France, whereas \cicero{}'s supply center count rises except playing as England. \cicero{} makes better strategic decisions.}
    \label{fig:faceted_scs}
\end{figure*}
\begin{table*}
    \centering
    \begin{tabular}{p{3cm}p{12cm}}
    \hline
    \textbf{Sender} & \textbf{Message}   \\ 
    \hline
     Germany & \small{This worked out great!} \\ 
      Germany & \small{Can we please talk about our moves a bit? It's very hard to coordinate with silence.} \\ 
    \rowcolor{grayish}  % Coloring a specific row
    France & \small{Absolutely! I'm all ears! What do you want to do now!} \\
        \rowcolor{grayish}  % Coloring a specific row
    France & \small{Any designs on Norway? I think you could get it this turn. I'm gonna go against England, as you see. Let's work together on England.} \\
        \rowcolor{grayish}  % Coloring a specific row
    France & \small{Come on mate, let's do better this turn and coordinate! What are your plans? I'm with you.} \\
        \rowcolor{grayish}  % Coloring a specific row
    France & \small{Alright, let's use the additional time! What do we do?} \\
         Germany & \small{I'm moving Sweden to Norway.} \\ 
      Germany & \small{Can we also start DMZing our border?} \\ 
              \rowcolor{grayish}  % Coloring a specific row
    France & \small{Nice, with support from Hel that should work out.} \\
    \rowcolor{grayish}  % Coloring a specific row
    France & \small{I'm not gonna move out of Belgium but I'll certainly not move any further either. I'm in against England. Can't fight both of you that's for sure.} \\
  Germany & \small{You should probably move Marseilles -> Spain.} \\ 
    \rowcolor{grayish}  % Coloring a specific row
    France & \small{Thank you! England might bring a fleet down? Good thought. Thank you!} \\
    \end{tabular}
    \caption{A conversation between France and Germany. They agree to DMZ (demilitarizing) their borders, e.g., Ruhr, and cooperate moves. However, Germany is deceptive and would rather move into Ruhr in this turn.}
    \label{fig:longconvo_dec}
\end{table*}
% \begin{figure*}
%     \centering
%     \includegraphics[width=0.9\textwidth]{figures/persuasion_by_turn.pdf}
%     \caption{Persuasion rate by turn. The rate is derived by persuasions by the number of live players (humans or \cicero{})}
%     \label{fig:persuasion_by_turn}
% \end{figure*}
% \begin{figure*}
%     \centering
%     \includegraphics[width=1.0\textwidth]{figures/long_conversasion_deception.pdf}
%     \caption{A conversation between France and Germany. They agree to DMZ (demilitarizing) their borders, e.g., Ruhr, and cooperate moves. However, Germany is deceptive and would rather move into Ruhr in this turn.}
%     \label{fig:longconvo_dec}
% \end{figure*}

\end{document}